\documentclass[10pt,twocolumn,letterpaper]{article}
\usepackage[accsupp]{axessibility} 
\usepackage{iccv}
\usepackage{times}
\usepackage{epsfig}
\usepackage{graphicx}
\usepackage{amsmath}
\usepackage{amssymb}

\usepackage{graphicx}
\usepackage{amsmath}
\DeclareMathOperator*{\argmax}{arg\,max} 
\DeclareMathOperator*{\argmin}{arg\,min} 
\usepackage{amssymb}
\usepackage{booktabs}
\usepackage{multirow}
\usepackage{algorithm}
\usepackage{algorithmic}
\usepackage{xcolor}
\usepackage{bm}
\usepackage[hyphens]{url}

\usepackage[pagebackref=true,breaklinks=true,letterpaper=true,color links,bookmarks=false]{hyperref}

\iccvfinalcopy 

\def\iccvPaperID{9264} 

\ificcvfinal\fi

\begin{document}

\title{Knowledge-Aware Federated Active Learning with Non-IID Data}

\author{Yu-Tong Cao$^{1}$, Ye Shi$^{2}$, Baosheng Yu$^{1}$,  Jingya Wang$^{2}$, Dacheng Tao$^{1}$ \\
$^1$ Sydney AI Centre, School of Computer Science, The University of Sydney \\
$^2$ ShanghaiTech University \\
{\tt\small ycao5602@uni.sydney.edu.au, shiye@shanghaitech.edu.cn, baosheng.yu@sydney.edu.au},  \\ {\tt\small wangjingya@shanghaitech.edu.cn, dacheng.tao@gmail.com}
}

\maketitle
\ificcvfinal\thispagestyle{empty}\fi

\begin{abstract}
Federated learning enables multiple decentralized clients to learn collaboratively without sharing local data. However, the expensive annotation cost on local clients remains an obstacle in utilizing local data. 
In this paper, we propose a federated active learning paradigm to efficiently learn a global model with a limited annotation budget while protecting data privacy in a decentralized learning manner. The main challenge faced by federated active learning is the mismatch between the active sampling goal of the global model on the server and that of the asynchronous local clients. This becomes even more significant when data is distributed non-IID across local clients. 
To address the aforementioned challenge, we propose Knowledge-Aware Federated Active Learning (KAFAL), which consists of Knowledge-Specialized Active Sampling (KSAS) and Knowledge-Compensatory Federated Update (KCFU). Specifically, KSAS is a novel active sampling method tailored for the federated active learning problem, aiming to deal with the mismatch challenge by sampling actively based on the discrepancies between local and global models. KSAS intensifies specialized knowledge in local clients, ensuring the sampled data is informative for both the local clients and the global model. Meanwhile, 
KCFU deals with the client heterogeneity caused by limited data and non-IID data distributions by compensating for each client's ability in weak classes with the assistance of the global model. 
Extensive experiments and analyses are conducted to show the superiority of KAFAL over recent state-of-the-art active learning methods.
Code is available at \url{https://github.com/ycao5602/KAFAL}.
\end{abstract}

\section{Introduction}
\begin{figure}[t]
    \centering
    \includegraphics[trim= 0 110 0 0,width=\linewidth, page=38]{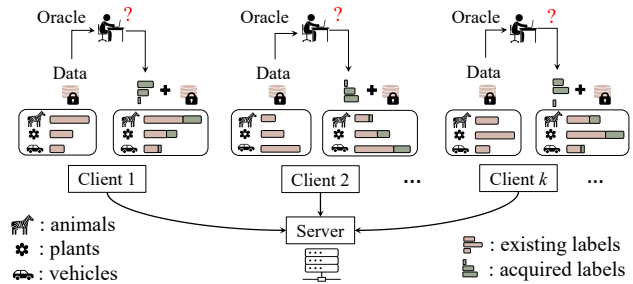}
    \caption{The primary federated active learning framework with non-IID data. Each client maintains an active learning loop to select informative data for annotation with a limited annotation budget. 
    We show each model's existing labelled data in different classes with pink bars and the newly acquired labels with green bars. Clients specialize in different classes due to non-IID data distributions.
    }
    \label{fig:focus}
\end{figure}

Federated learning is a decentralized paradigm that allows collaborative learning of local devices to attain a powerful global model in a central server through aggregation without accessing local data~\cite{konevcny2016federated,mcmahan2017communication}. Most federated learning methods consider supervised learning scenarios with fully annotated training data on each local client. However, the high annotation cost has been a challenge for real-world federated learning  scenarios, e.g., large-scale medical data located in different medical institutions while medical specialists for data annotation are very limited in each institution. In this paper, we consider a new federated active learning paradigm, which aims to not only protect data privacy but also make the most of the very limited annotation budget on each local client for decentralized model training. An illustration of the proposed federated active learning with non-IID data framework is shown in Fig.~\ref{fig:focus}. 

In federated active learning, we aim to attain a powerful global model on the server by sampling only local data and training the model on each local client. 
A straightforward solution for federated learning is to directly apply off-the-shelf active learning methods to each client. Specifically, existing methods can mainly be categorized into diversity-based~\cite{sener2017active,ash2020deep}, uncertainty-based~\cite{yoo2019learning,sinha2019variational,zhang2020state,wangdual,kim2021task,ebrahimi2020minimax,beluch2018power}, and discrepancy-based. \cite{seung1992query,cortes2019active}. Therefore, we actively sample based on either the statistics from each client model or the downloaded global model. However, the former approach may yield benefits primarily for local clients, while the latter might result in the loss of valuable information during aggregation, even if the selected data is advantageous for the global model on the server. 
Through experiments in subsequent sections, we demonstrate that active sampling with the global model on the server struggles to derive benefits due to this indirect process.

A major challenge in federated active learning is therefore, the mismatch between the active sampling goal of the clients and that of the model on server caused by asynchronous models. What makes it even more challenging is the statistical heterogeneity resulting from the non-IID data distributions on clients in a typical federated learning setting~\cite{mcmahan2017communication,zhao2018federated,hsieh2020non,li2020federated}. 
Ideally, the models can synchronize with sufficiently many aggregations from local clients to the model on the server. However, the communication costs usually make the above-mentioned solution impractical~\cite{mcmahan2017communication}. Therefore, the model parameters of each client and the global model vary due to non-IID distributions, leading to a higher degree of mismatch between the sampling goals.

To address the aforementioned challenge, we propose a federated active learning scheme, namely  {\bf Knowledge-Aware Federated Active Learning (KAFAL)}. It comprises two key components, Knowledge-Specialized Active Sampling and Knowledge-Compensatory Federated Update.
{\bf Knowledge-Specialized Active Sampling (KSAS)} is a new active sampling strategy, where each client model learns to intensify its specialized knowledge in order to annotate universally informative data that benefit both the clients and the global model.
Specifically, we compute the intensified discrepancy between the client and global model outputs based on the specialized knowledge of each client. 
In addition, the insufficiency of labelled training data together with the statistical heterogeneity caused by non-IID data can degrade the federated update quality, e.g., clients may perform weakly for certain classes. Aggregating these clients, extra communications are required to achieve convergence. Therefore, we further devise a new update rule, {\bf Knowledge-Compensatory Federated Update (KCFU)}, by compensating for weak classes (or low-frequency classes) on each client through knowledge distillation from the global model.  The main contributions of this paper are as follows: 
 
\begin{itemize}
 \item  We explore a rarely studied problem, federated active learning with non-IID data, which aims at efficiently learning a global model with a limited annotation budget on each client under a heterogeneous federated learning framework. Notably, we reveal the main challenge in federated active learning is the mismatch between the active sampling goal of the clients and that of the server caused by asynchronous models. 
 \item We introduce a federated active learning paradigm, known as KAFAL, with a novel active sampling method KSAS and a novel federated update method KCFU to handle the aforementioned challenge. KSAS is designed to sample universally informative data by computing the intensified discrepancies between the clients' and the global model's outputs based on the specialized knowledge of each client. KCFU is devised to deal with data heterogeneity by compensating for weak classes using knowledge distillation from the global model. 
 \item We conduct extensive experiments on different benchmarks to demonstrate the superiority of the proposed method, where comprehensive ablation studies are also provided to validate the design of the proposed KAFAL.  
\end{itemize}

\section{Related Work}
\subsection{Federated Learning}
Federated learning is a learning paradigm that allows decentralized training of a model on the central server with training data distributed over a number of local clients in a non-IID manner~\cite{konevcny2016federated,mcmahan2017communication,hsu2019measuring,mohri2019agnostic,hsu2020federated,chen2021bridging,lin2020ensemble,gong2021ensemble}. Specifically, Konevcny et al.~\cite{konevcny2016federated} first introduced the term and proposed a method, FedAvg, to aggregate the client models, which was later improved by FedAvgM to accumulate model updates with momentum~\cite{hsu2019measuring,hsu2020federated}. 
Federated learning has also been discussed in more practical views, such as federated multi-task learning~\cite{marfoq2021federated}, federated domain adaptation~\cite{peng2019federated,yao2022federated}, federated continual learning~\cite{yoon2021federated}, semi-supervised federated learning~\cite{jeong2020federated,wang2021federated}, and unsupervised federated learning~\cite{lu2021unsupervised}. Specifically,
Jeong et al.~\cite{jeong2020federated} considered the deficiency of data labels in federated learning and proposed a semi-supervised solution. 
Ahn et al.~\cite{ahn2022federated} and Kim et al.~\cite{kimlg} discussed a federated active learning paradigm, while they only considered the less realistic IID data scenario. 
To the best of our knowledge, we are the first to explore the active data sampling problem in the non-IID federated learning framework.  

\subsection{Active Learning}
Existing active learning methods can be categorized into diversity-based, uncertainty-based, and discrepancy-based methods. Specifically,
diversity-based methods~\cite{sener2017active,ash2020deep, parvaneh2022active} select representative and diverse data points that span the data space for query. Sener et al.~\cite{sener2017active} proposed a core-set approach that selects the most representative core-set from the data pool using k-center algorithms. Recently, Ash et al.~\cite{ash2020deep} proposed to actively select the data points that produce gradients with diverse directions. 
Uncertainty-based methods~\cite{yoo2019learning,sinha2019variational,zhang2020state,wangdual,kim2021task,beluch2018power} estimate the uncertainty of predictions using different metrics and select data points accordingly. 
Despite being simple to use, these methods cannot be directly applied in federated active learning, without the mismatch between each client model and the global model being handled.
Some recent methods explicitly measure the informativeness of data points instead of directly calculating the uncertainty metrics~\cite{yoo2019learning,sinha2019variational,zhang2020state,wangdual,ebrahimi2020minimax}. Specifically, Sinha et al.~\cite{sinha2019variational} utilized an extra variational auto-encoder to select data points that are less likely to be distributed in the labelled pool for querying. Despite effective sampling, these methods require extra modules for sampling with an increased computational cost.
Discrepancy-based methods~\cite{seung1992query,freund1993information,dagan1995committee,melville2004diverse,cortes2019active} pass data points through an ensemble of models, namely a committee, and select the data points that cause large discrepancy within the committee. Freund et al.~\cite{freund1993information} proposed to randomly pick two models in the committee that are consistent for labelled data and then use them to sample from unlabelled data.  
Multiple models usually make discrepancy-based methods stable, but also increase the computational cost. This partially explains why they become less popular with the rise of deep active learning. It is costly to fit them in federated active learning.

Many recent methods have also been proposed to enable active learning in more challenging settings, e.g., low-budget active learning~\cite{mahmood2021low}, biased-data active learning~\cite{gudovskiy2020deep}, semi-supervised active learning~\cite{gao2020consistency,huang2021semi} and cross-domain active learning~\cite{fu2021transferable,ma2021active}. 
Our work also considers applying active learning in a more practical decentralized federated learning setting where local data privacy is protected. Chen et al.~\cite{chen2020autodal} proposed a novel automated learning system for distributed active learning that requires a shared labelled set. Furthermore, Goetz et al.~\cite{goetz2019active} considered active learning in a federated learning framework that studies how to select clients actively. Our work, instead, considers the active sampling of data on each local client in federated learning. 

\section{Method}

In this section, we first describe the problem setting of federated active learning and then introduce two main components, i.e., KSAS and KCFU in the proposed KAFAL.

\begin{algorithm}[tb]
\caption{Knowledge-Aware Federated Active Learning}
\label{alg:KAFAL}
\textbf{Data}: local datasets $\{\mathcal{D}_k^L\}^{K}_{i=1}$ and $\{\mathcal{D}_i^U\}^{K}_{i=1}$ \\
\textbf{Input}: $T$, $R$, sampling budgets $\{b_i\}_{i=1}^K$  \\
\textbf{Parameter}: $\bm\Omega$, $\{\bm\omega_i\}^{K}_{i=1}$
\begin{algorithmic}[1] 
\FOR{active round a=1 to A}
\STATE{\bf\underline{Federated Update: KCFU}}
\STATE{Initialize the global model with $\bm\Omega^0$}
\FOR{communication round $t=1$ to $T$}
    \STATE{$S_t\leftarrow$ Random subset of $\lceil R \cdot K\rceil$ clients.}

    \FOR{client $k$ $\in$ $S_t$}

        \STATE{Download the global model's parameters $\bm\Omega^t$ }
        
        \STATE{Copy to the client ${\bm\omega}_k^{t} \leftarrow\bm{{\Omega}}^t$}
        
        \STATE{${\bm\omega}_k^{t+1} \leftarrow$ LocalUpdate(${\bm\omega}_k^t; \mathcal{D}_k^L,\mathcal{D}_k^U$)}
        
        \STATE{Upload the local model parameters to the server}
    \ENDFOR
    \FOR{client $k'$ $\notin$ $S_t$}
        \STATE{Keep the client model unchanged ${\bm\omega}_{k'}^{t+1} \leftarrow{\bm\omega}_{k'}^{t} $}
    \ENDFOR
    \STATE{Aggregate the clients with Eq. (7) to update $\bm{\Omega}^{t+1}$ }
    
    \FOR{each client $k\in S_t$}
        \STATE{Download $\bm{\Omega}_k^{t+1}$ and save as $\bm{\hat\Omega}_k$} 
    \ENDFOR
\ENDFOR
\STATE{\bf\underline{Active Sampling: KSAS}}
\FOR{client $i=1$ to $K$}
    \FOR{each unlabelled data $x\in\mathcal{D}^U_i$}
        \FOR{class $y\in \mathbb{C}$}
            \STATE{Compute $P_y^i(\bm x)$ on class $y$ using Eq. (1)}

            \STATE{Compute $Q_y^i(\bm x)$  on class $y$ using Eq. (2)}
        \ENDFOR
    
    \STATE{Compute $D^i(\bm x)$ using Eq. (3)}
    \ENDFOR
    \STATE{Send $b_i$ unlabelled data points with the largest $D$ to the oracle for annotation}
    
    \STATE{Remove the annotated data in $\mathcal{D}^U_i$ and add to $\mathcal{D}^L_i$}
\ENDFOR    
\ENDFOR  
\STATE \textbf{Return} $\{\mathcal{D}_i^L\}^{K}_{i=1}$ and $\{\mathcal{D}_i^U\}^{K}_{i=1}$ 
\end{algorithmic}
\end{algorithm}

\subsection{Problem Setting}

We illustrate the overview of the federated active learning framework in Fig.~\ref{fig:focus} and sum up the proposed KAFAL algorithm in Alg.~\ref{alg:KAFAL}. In federated active learning, we keep $K$ local client models parametrized with $\{{\bm \omega}_{i}\}^{K}_{i=1}$ and one global model on central server parametrized with $\mathbf\Omega$. Each client model $i$ is optimized using its local training dataset $\mathcal{D}_{i}$. 
Different from standard federated learning, federated active learning annotates a subset of data samples on each client with a local active learning loop. The training set $\mathcal{D}_{i}$ for client $i$ is divided into a labelled set $\mathcal{D}^{L}_{i}$ and an unlabelled set $\mathcal{D}^{U}_{i}$. 
In each communication round, a fraction $R\in(0,1]$ of the total $K$ clients are first randomly selected as a subset $S_t$, which simulates the real-world scenarios that some local devices may be offline from time to time.
After that, the selected clients first download $\mathbf\Omega$ from the server to initialize $\{{\bm \omega}_{k}\}_{k\in S_t}$, and then conduct local update based on $\{\mathcal{D}_{k}\}_{k\in S_t}$. The updated $\{{\bm \omega}_{k}\}_{k\in S_t}$ will be uploaded to the central server and aggregated to update ${\mathbf\Omega}$. The training process terminates after $T$ communication rounds. 
After that, a batch of unlabelled data is sampled from each $\mathcal{D}^{U}_{i}$, sent to the local oracle for annotation, and added to the labelled data pool $\mathcal{D}^{L}_{i}$ for each client $i$. The sampling budget for the client $i$ is $b_i$. The active sampling process is repeated for $A$ times, where $A$ is set according to need. 
The training sets $\{\mathcal{D}_{i}\}^{K}_{i=1}$ follow non-IID distributions. All client models share the same architecture with the global model to synchronize model parameters between the client and server.
Just like in federated learning, transferring the local data $\{\mathcal{D}_{i}\}^{K}_{i=1}$ across clients (or server) is prohibited in federated active learning. 
The objective of federated active learning is to actively annotate local data with limited budgets to improve the overall model performance without violating data privacy. 
 
{
\subsection{Knowledge-Specialized Active Sampling}
Given the mismatch problem in federated active learning, informative data on each client may not be that informative to the global model due to the non-IID data distributions, meaning that using only one of them for active sampling is therefore not reliable.
Computing the model discrepancy between each client and the global model allows us to consider both aspects in active sampling. 
But alone is insufficient.
Data from rare classes in each local dataset can cause large discrepancies between the client and the global model. However, they are usually uninformative to the global model and can hardly contribute to the client model's updates. Being rare locally makes their contributions limited in the gradient computation. Furthermore, during aggregation, the global model may not find them as informative as they are to the clients.
Hence, we propose to enable each client to intensify its specialized knowledge (common class knowledge) in the computation of discrepancy to sample more informative data containing specialized knowledge. We introduce the Knowledge-Specialized KL-Divergence as follows. On top of a symmetrized KL-Divergence~\cite{jeffreys1939theory,kullback1951information}, our Knowledge-Specialized KL-Divergence further incorporates a Knowledge-Specialized component to accentuate each client's specialized knowledge. 
The Knowledge-Specialized probability of client $i$ being predicted to class $y$ is formulated as:
\begin{equation}
P_{y}^i(\bm x) =  \frac{n_{i,y}^{\lambda}\text{exp}\Big( g_y({\bm x};{\bm\omega_i})\Big)}{\sum_{c\in\mathbb{C}}n_{i,c}^{\lambda}\text{exp}\Big(g_c({\bm x};{\bm\omega_i})\Big)},
\label{eq:ks-prob-p}
\end{equation}
where $\bm x$ is an unlabelled data point sampled from $\mathcal{D}^U_i$, $g_y({\bm x};{\bm\omega_i})$ is the prediction score at the $y$-th class, $\mathbb{C}$ indicates the set of all classes, $n_{i,y}$ is the number of data points that belong to class $y$ in $\mathcal{D}_i^L$, and $\lambda$ is a hyperparameter which controls the knowledge-specialized level. We name $n_{i,y}^\lambda$ as the knowledge weight which indicates the client's knowledge in each class.
Similarly, the knowledge-specialized probability of the global model predicted to be class $y$ can be defined as:
\begin{equation}
Q_y^i(\bm x) =  \frac{n_{i,y}^{\lambda}\text{exp}\Big( g_y({\bm x};\mathbf{\hat\Omega}_i)\Big)}{\sum_{c\in\mathbb{C}}n_{i,c}^{\lambda}\text{exp}\Big(g_c({\bm x};\mathbf{\hat\Omega}_i)\Big)},
\label{eq:ks-prob-q}
\end{equation}
where $\mathbf{\hat\Omega}_i$ is a copy of global model parameters downloaded from the server to client $i$. 
The knowledge-specialized KL-Divergence is defined as:
\begin{equation}
\small
D^i(\bm x)=\sum_{y\in\mathbb{C}}\Big( P_y^i(\bm x)\ln \frac{P_y^i(\bm x)}{Q_y^i(\bm x)} + Q_y^i(\bm x)\ln \frac{Q_y^i(\bm x)}{P_y^i(\bm x)} \Big), 
\label{eq:ks-kld}
\end{equation}
where $\bm x$ is data from the unlabelled pool of client $i$. The knowledge-specialized KL-Divergence focuses on each client's specialized knowledge and selects more informative data points from its specialized classes for labelling. 
In the Knowledge-Specialized probabilities, $\{n_{i,c}\}_{c\in\mathbb{C}}$ serve to amplify the KL-Divergence on class $c$ if the class is considered to contain the client's specialized knowledge. 
\begin{figure}[t]
    \centering
    \includegraphics[trim=0 140 0 0,clip,width=\linewidth, page=56]{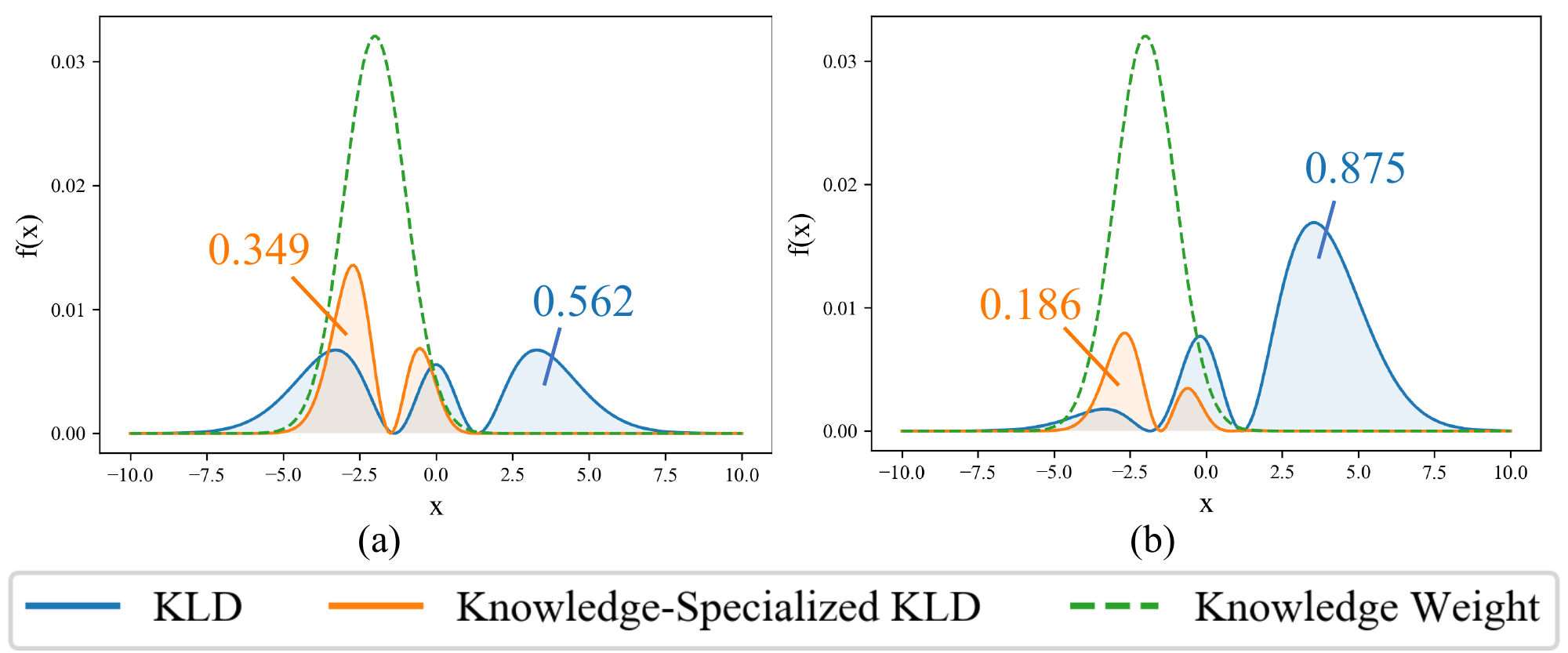}
    \caption{
    Illustration of how Knowledge-Specialized KL-Divergence intensifies specialized knowledge compared to the standard KL-Divergence. The blue and orange lines integrate to be KL-Divergence and the knowledge-specialized KL-Divergence computed from the same pair of distributions. The blue and orange numbers show the integrated areas of the blue and orange curves in each image, respectively. 
    }
    \label{fig:ksas-vis}
\end{figure}
{
\paragraph{Visualization.}
To better visualize how Knowledge-Specialized KL-Divergence intensifies specialized knowledge compared to the standard KL-Divergence, we use continuous distributions to simulate model predictions and compute the divergences in Fig.~\ref{fig:ksas-vis}. Note that the knowledge weight curves represent a continuous version of our knowledge weights. For clarity, we only show the KLD and Knowledge-Specialized KLD and omit the distribution curves in the figure. (a) and (b) can be viewed as global-local discrepancies from two different inputs on the same client model since the KLD values are different and the knowledge weights are the same. Although (a) has a smaller KLD, its knowledge-specialized KLD is larger, meaning that if we used KLD for sampling, (a) is less likely to be sampled. On the other hand, if our proposed knowledge-specialized KLD is used, (a) is more likely to be sampled than (b). What makes the results different is the knowledge weight. It intensifies the client's specialized knowledge and suppresses the less reliable divergence contributed by unfamiliar knowledge of the client. More of the model difference in (a) is caused by specialized knowledge (peak area of knowledge weight) other than that in (b). More analyses are provided in the supplementary (\ref{supp:a10}).}

\subsection{Knowledge-Compensatory Federated Update}
\begin{figure}[t]
    \centering
    \includegraphics[trim=0 0 300 0,clip,width=\linewidth, page=58]{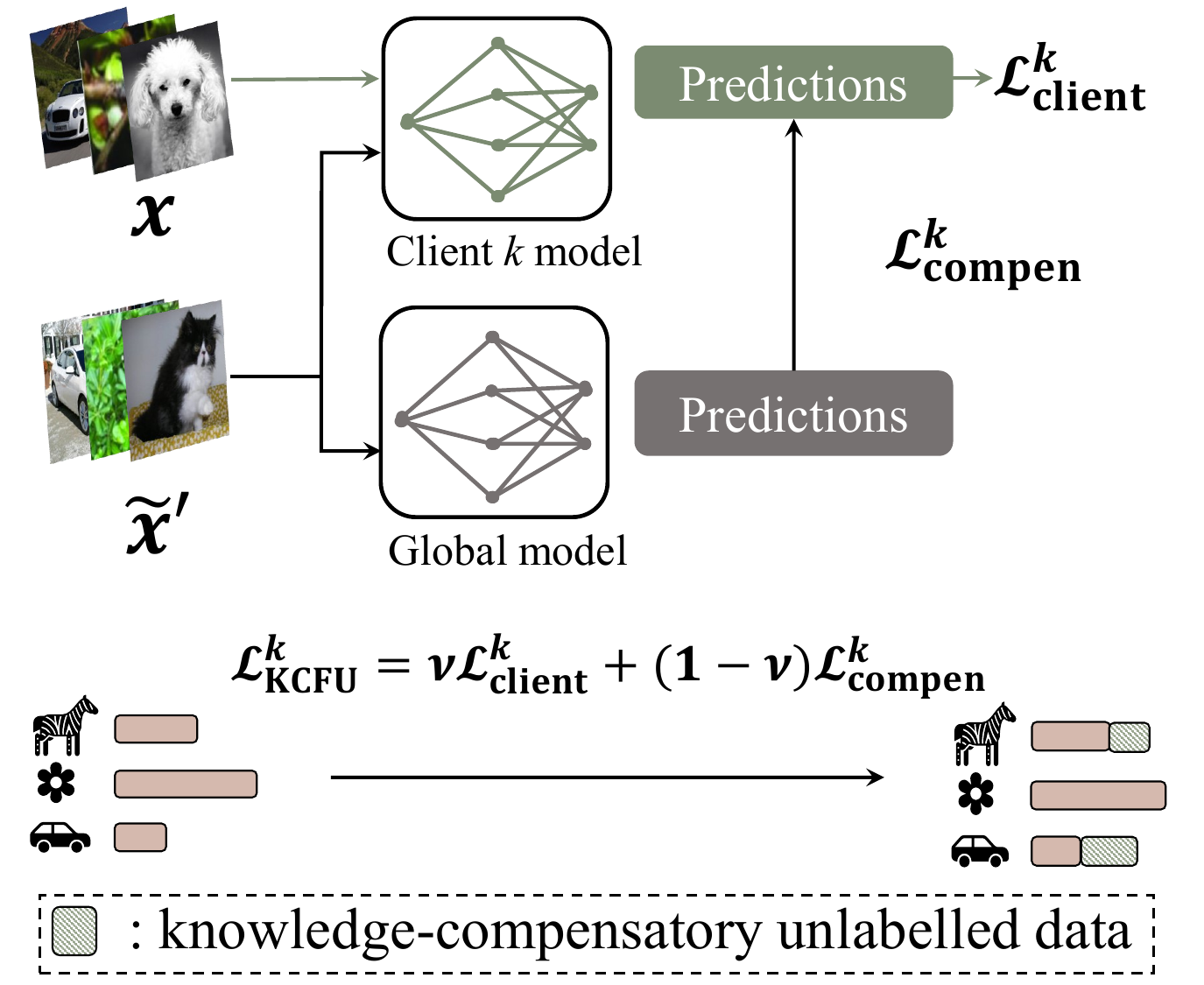}
    \caption{
    KCFU compensates for each client's ability on weak classes through knowledge distillation from the global model. Unlabelled data are used in the process.
    }
    \label{fig:kcfu}
\end{figure}
An overview of knowledge-compensatory federated update (KCFU) is shown in Fig.~\ref{fig:kcfu}. 
The local data on each client follows its own realistic data distributions~\cite{hsu2019measuring}, thus leaving non-uniform class distribution on each client. Besides, our KSAS which tends to annotate data with specialized-knowledge further introduces imbalance in labelled data. 
Therefore, on top of the standard FedAvg~\cite{mcmahan2017communication}, we introduce a balanced classifier and a knowledge-compensatory strategy.
\paragraph{Local Update with Balanced Loss.} Our balanced classifier on each client optimizes with a balanced cross-entropy loss~\cite{ren2020balanced} to deal with the imbalanced local data distribution:
\begin{equation}
\mathcal{L}_{\text{client}}^k=-\text{log}\frac{n_{k,y}\text{exp}\Big( g_y({\bm x};{\bm\omega_k})\Big)}{\sum_{c\in\mathbb{C}}n_{k,c}\text{exp}\Big(g_c({\bm x};{\bm\omega_k})\Big)}.
\label{eq:g-loss}
\end{equation}
In our experiment, each labelled set $\mathcal{D}_k^L$ starts from only a small proportion of the local dataset on the client $k$. We demonstrate with experiments in later sections that, with the imbalanced class distributions and the small size of training data, a simple local client update using cross-entropy loss is not enough for training. 
The balanced loss allocates more weight to data from rare classes and less weight to data from common classes to deal with the problem. It prevents the model from becoming biased towards common classes during training. 

\noindent\textbf{\textit{Remark:}} Although our balanced loss (Eq.~\eqref{eq:g-loss}) looks similar in formulation compared with the aforementioned knowledge-specialized probabilities, i.e., Eq.~\eqref{eq:ks-prob-p} and \eqref{eq:ks-prob-q}, they are designed for various purposes and function differently.
Eq.~\eqref{eq:ks-prob-p} and \eqref{eq:ks-prob-q} are designed to compute the active sampling scores, and
Eq.~\eqref{eq:g-loss} is a loss that updates the client models. The knowledge-specialized probabilities magnify the KL-Divergence computed from common classes for sampling, while the balanced loss
magnifies the loss computed from rare classes. 

\paragraph{Global-to-Local Knowledge Compensation.}
Due to the extreme limitation and non-uniformity of local data labels, the clients can perform weakly on rare classes. The weak classes of clients differ depending on the data distributions. 
Such statistical heterogeneity of clients can be harmful in model aggregation. 
To compensate for the clients' knowledge on the weak classes, we further introduce an extra loss $\mathcal{L}_\text{compen}$.
Since the global model aggregates parameters of local clients, they usually have a more balanced performance over different classes. On classes where each client considers to be rare, the global model is likely to perform better than the client. Hence, it is reasonable to design the knowledge-compensation process which conducts knowledge distillation from the global model to the clients using unlabelled data.  
We later show with experiments that $\mathcal{L}_\text{compen}$ can save the communication cost via boosting the convergence.
The loss $\mathcal{L}_\text{compen}$ on client $k$ can be evaluated as follows. We first sample unlabelled data ${\bm x'}$ from $\mathcal{D}_k^U$. Then we compute the logits ${\bm z} = g({\bm x'};{\bm{\Omega}})$ and the pseudo label ${y'} = \argmax\limits_c g_c({\bm x'};{\bm\Omega})$ with the downloaded global model. 
The loss weight can be computed as $\bm\Gamma(\bm x')=\frac{\sum_{c\in\mathbb{C}}n_{k,c}}{n_{k,y'}}$, 
and the compensation loss is then defined as:
\begin{equation}
\begin{split}
\mathcal{L}_\text{compen}^k= {\bm\Gamma(\bm x')}\cdot \text{KL}\Big(\sigma({\bm z})\| \sigma({g({\bm x'};{\bm\omega_k})})\Big) ,
\label{eq:compen-loss}
\end{split}
\end{equation}
where $\sigma$ stands for the softmax function and KL-divergence $\text{KL}(p,q) = p\ln\frac{p}{q}$. Note that no gradient is computed for the global model $\Omega$, only. As the unlabelled data falls in the same distribution as the labelled data, rare classes in labelled data are usually still rare in unlabelled data. To make the most of the compensation loss, we further propose to augment the training with mixed unlabelled data $\bm{\tilde{x}'} = {\bm\beta} {\bm x'_1 + ({{\bm 1}}-{\bm\beta}){\bm x'_2}}$, where ${\bm\beta}$ is a mixing weight sampled from a beta distribution. $\bm x'_1$ and ${\bm x'}_2$ are randomly sampled from the unlabelled batch. $\Gamma(\bm{\tilde{x}'})$ is similarly mixed as ${\bm\beta} {\Gamma({\bm x'_1}) + ({{\bm 1}}-{\bm\beta}){\Gamma({\bm x'_2})}}$. The compensation loss then becomes $\mathcal{L}_\text{compen}(\bm{\tilde{x}'}; {\bm {\tilde{z}}},{\bm\omega_k})$ with $\bm {\tilde{z}} = g({\bm{\tilde{x}'}};{\bm{\Omega}})$. 
Therefore, the complete loss $\mathcal{L}^k_\text{KCFU}$ to update client $k$ is:
\begin{equation}
\begin{split}
\mathcal{L}_\text{KCFU}^k&=\nu\mathcal{L}_\text{client}^k + (1-\nu)\mathcal{L}_\text{compen}^k, 
\label{eq:KCFU-loss2}
\end{split}
\end{equation}
where $\nu$ is a tradeoff hyperparameter.
We show the detailed local update algorithm in Alg.~\ref{alg:local}. 

\begin{algorithm}[tb]
\caption{LocalUpdate$({\bm\omega_k};\mathcal{D}_k^L,\mathcal{D}_k^U)$}
\label{alg:local}
\textbf{Data}: $\mathcal{D}_k^L$, $\mathcal{D}_k^U$\\
\textbf{Input}: epochs, batches, communication round $t$, learning rate $\eta$ \\
\textbf{Parameter}: $\bm\omega_k$
\begin{algorithmic}[1] 
\FOR{$e=1$ to epochs}
    \FOR{$b=1$ to batches} 
      \STATE{Sample a batch $\{(\bm{x},y)\}\subseteq\mathcal{D}_k^L$} 
      \STATE{Compute $\mathcal{L}_\text{client}^k$ using Eq. (4) }
      \IF{$t$ equals 1}
          \STATE{${\bm\omega_k}\leftarrow{\bm\omega_k}-\eta\nabla\mathcal{L}_\text{client}^k$}
      \ELSE
          \STATE{Sample a batch $\{\bm{x'}\}\subseteq\mathcal{D}_k^U$}
          \STATE{Construct a mixed batch $\{\bm{\tilde{x}'}\}$}
          \STATE{Find ${\bm z}$, $\bm\Gamma(\bm{\tilde{x}'})$ for each $\bm{\tilde{x}'}$}
          \STATE{Compute $\mathcal{L}_\text{ compen}^k$ using Eq. (5) }
          \STATE{Compute $\mathcal{L}_\text{KCFU}^k$ with Eq. (6)}
          \STATE{${\bm\omega_k}\leftarrow{\bm\omega_k}-\eta\nabla\mathcal{L}_\text{KCFU}^k$}
      \ENDIF
    \ENDFOR
\ENDFOR
\STATE \textbf{Return} $\bm\omega_k$
\end{algorithmic}
\end{algorithm}

\paragraph{Global Aggregation} After local updates of clients, they are uploaded to the server and aggregated as follows:
\begin{equation}
\mathbf{\Omega}^{t} =\sum_{k\in S_t}\frac{N_k}{\sum_{j\in S_t}N_j}{\bm\omega}^{t}_k,
\label{eq:aggregation}
\end{equation}
where $N_k=\sum_{c\in\mathbb{C}}n_{k,c}$ indicates the number of data points in local labelled data pool $\mathcal{D}_k^{L}$. 
Lastly, we can formulate the overall objective as:
\begin{equation}
\begin{split}
 \argmin_{\{\bm\omega_k\}_{k\in S_t},\bm x\sim\{\mathcal{D}_k^L\}_{k\in S_t}, \bm x'\sim\{\mathcal{D}_k^U\}_{k\in S_t}}\mathcal{L}_\text{KCFU}, 
\label{eq:KCFU-loss-total}
\end{split}
\end{equation}
where $\mathcal{L}_\text{KCFU}=\sum_{k\in S_t}\frac{N_k}{\sum_{j\in S_t}N_j}\mathcal{L}_\text{KCFU}^k$, $\{\mathcal{D}_k^L\}_{k\in S_t}$ is achieved via active learning loops. 

}

\begin{figure*}[t]
    \centering
    \includegraphics[trim= 0 70 0 0, width=\textwidth, page=36]{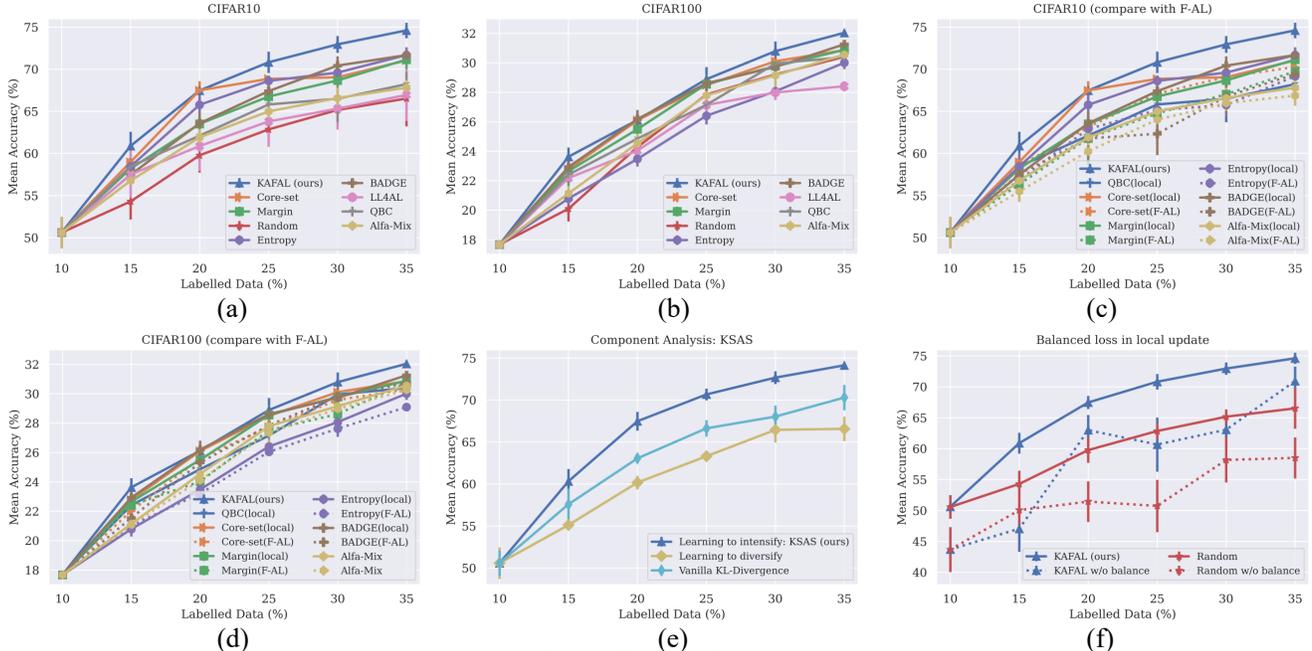}
    \caption{(a)-(b) The federated active learning results from different active learning baselines plus the results of our KAFAL on CIFAR10/100 with $\alpha=0.1$. (c)-(d) Comparing our KAFAL with the federated active learning baseline F-AL on CIFAR10/100 with $\alpha=0.1$. (e) Component analyses of KSAS and KCFU on CIFAR10. (f) Results with balanced loss (Eq.~\eqref{eq:g-loss}) and standard cross-entropy loss on CIFAR10. For all figures, the error bars show the standard deviation of results across $5$ runs.
    }
    \label{fig:experiments}
\end{figure*}

\section{Experiments}
In this section, we mainly conduct experiments on three image classification datasets, CIFAR10/100~\cite{krizhevsky2009learning} and MNIST~\cite{lecun2010mnist}, that are popular in both active and federated learning. Additionally, we also apply our method in a more realistic scenario by conducting medical image classification with NIH Chest X-Ray dataset~\cite{wang2017hospital}. Specifically, 
CIFAR10 and CIFAR100 contain $60,000$ images from $10$ and $100$ classes, respectively, including $50,000$ training images and $10,000$ testing images. Results and details of MNIST are shown in the supplementary (\ref{supp:a8}). For the server and client models, we utilize ResNet-8~\cite{he2016identity} as the model architecture. 

We implement the method with Pytorch~\cite{paszke2017automatic}. We use $K=10$ clients in our experiments. In each communication round, $R=80\%$ of the clients are selected at random to update locally. The hyperparameter $\lambda$ is set to $1$. More details are given in the supplementary (\ref{supp:a1}).
To distribute non-IID data to different clients, we follow Hsu et al.~\cite{hsu2019measuring} and draw $q \sim \text{Dir}(\alpha p)$ from a Dirichlet distribution. p stands for the global prior class distribution over all classes, and $\alpha > 0$ is a concentration parameter that controls the level of IID.
When $\alpha\rightarrow\infty$, the data distributions are identical to the global class distribution. When $\alpha\rightarrow 0$, each client will be allocated data from only one class. In our main results, we set $\alpha=0.1$. We also show the results from $\alpha=0.3$ and $\alpha=1$. The different CIFAR10 data distributions are shown in the supplementary (\ref{supp:a1}). 

For active learning loops, we start by randomly selecting $10\%$ data from $\mathcal{D}_i$ as the labelled pool $\mathcal{D}^L_i$ of client $i$. This is around $500$ labelled data for each client. For each sampling cycle, the budget $b_i$ on each client is $5\%$ of the total local data $\mathcal{D}_i$. We sample for $A=5$ times until the labelled data amount reaches $35\%$ of all data for each client. We repeat each experiment $5$ times with different random seeds and average the results to get a final result. 

\subsection{Comparison with Active Learning Methods}

\label{sec:sota}
We compare our KAFAL with $8$ other active learning methods and show the results in Fig.~\ref{fig:experiments}(a)(b). All methods are fit into the federated active learning framework using the same model architectures following the same training steps for fair comparison. For all baselines, we use the KCFU loss (Eq.~\eqref{eq:KCFU-loss2}) for local update. FedAvg is used for aggregation for all methods. We categorize our baselines into five types. (I) We compare with uncertainty-based methods: entropy and top-2 margin scores (Margin). Entropy is calculated as $H(p) = -p\cdot \text{log}(p)$, where $p$ is the Softmax output. The top-2 margin score calculates the margin between the largest prediction score and the second largest prediction score over all classes for each data point. Unlabelled data with the lowest top-2 margin scores will be sampled for annotation. Here we compute the uncertainty scores on each client model after local update. (II) We compare with a special uncertainty-based method which explicitly learns the data loss with extra modules, Learning Loss for Active Learning (LL4AL)~\cite{yoo2019learning}. We train a loss prediction module for each client model. (III) We also compare with diversity methods: Core-set~\cite{sener2017active}, BADGE~\cite{ash2020deep}, and ALFA-Mix~\cite{parvaneh2022active}. We sample on each client model using diversity.
(IV) Results from a previous discrepancy-based method, Query-by-Committee (QBC)~\cite{dagan1995committee}, is also compared with. We use $3$ models on each client for QBC. (V) Finally, we compare with random sampling results.  

On both CIFAR10 and CIFAR100, our KAFAL achieves state-of-the-art results (Fig.~\ref{fig:experiments}(a)(b)). 
The margins between KAFAL and other baselines become larger with the increase of labelled data.
On CIFAR10, our method eventually achieve a margin of around $3\%$ compared to BADGE, Entropy, Margin, and Core-set. LL4AL, although quite competitive in standard active learning, does not perform well in the federated active setting. Besides, LL4AL and QBC update extra model parameters of sizes $0.015M$ and $0.156M$ for each client, when each client's model size for the rest methods is only $0.078M$. On CIFAR100, the margins are less significant compared to results from CIFAR10.  
Probably because the $10$ times of classes in CIFAR100 makes it a much more difficult task, especially consider the limited amount of labelled data for each client. Some of the methods perform poorer than Random, possibly because Random naturally diversifies in sampling. It is worth noting that in CIFAR10 and CIFAR100, the full-set federated learning results are $72.93\%$ and $37.35\%$. On CIFAR10, the full-set result is lower than our KAFAL result with $35\%$ labelled data. This could be because our strategy selects only the most informative data for annotation and avoids data redundancy. On CIFAR100, the full-set result is around $5\%$ higher than our KAFAL result with $35\%$ labelled data, indicating that $35\%$ data is not enough to represent a $100$-class dataset. Notably, we also evaluate each client with the test set and show the analysis in the supplementary (\ref{supp:a1}) for a complete picture of the performance of our KAFAL.

{
\subsection{Comparison with Sampling by Global Model} 
As we mentioned in previous sections, for some sampling methods, it is possible to compute the sampling criteria either on the local client after local updates or on the downloaded aggregated global model. Using the global model for sampling is also the main idea of F-AL~\cite{ahn2022federated}, a federated active learning method for IID data. Among the baselines we compared with in Sec.~\ref{sec:sota}, we found Core-Set, Margin, Entropy, BADGE, and Alfa-Mix to be qualified to compute sampling scores on either the clients or the downloaded global model. We show the experiment results on CIFAR10 and CIFAR100 in Fig.~\ref{fig:experiments}(c)(d). The solid lines show the results from using locally updated client models to compute sampling criteria. These are also the results presented in Fig.~\ref{fig:experiments}(a). The dashed lines represent the results from using the downloaded global model for computing sampling scores. There is a clear drop in performance moving from client model statistics to global model statistics.  Additionally, we compare our method with QBC. Our method combines local and global with discrepancy-based sampling and QBC is a local disagreement-sampling method. 
This experiment demonstrates the challenge in federated active learning where the sampling aims of the clients mismatch with that of the global model. It also shows that even if we sample informative data points directly using the downloaded global model, the information cannot be fully utilized to benefit the global model through aggregation. F-AL, which is initially proposed for IID federated active learning, does not suit the task of federated active learning with non-IID data.
}

\subsection{Ablation Studies}
\subsubsection{Component study}
To explore the importance of our model components in KAFAL, we separately run experiments to analyze KASA and KCFU. 
To analyze KSAS, we first replace our knowledge-Specialized KL-Divergence with a vanilla KL-Divergence and observe a $3\%$ to $5\%$ performance drop through the whole sampling process (Fig.~\ref{fig:experiments}(c)). We also attempt to sample with a reversed KSAS (learning to diversify), where we replace each $n_{i,y}$ and $n_{i,c}$ in Eq.~\eqref{eq:ks-prob-p} and \eqref{eq:ks-prob-q} with $\frac{1}{n_{i,y}}$ and $\frac{1}{n_{i,c}}$. This prevents the client models from intensifying their specialized knowledge. Instead, it drives the clients to focus on sampling data from rare classes. The results show a significant drop compared to the other two. This further validates our design where each client should intensify its knowledge during active sampling. Data from rare classes can be quite useless in improving the global model on the server. 

To analyze the efficiency of KCFU, we count the number of communication rounds of different federated update ways to achieve the same accuracy. The benchmark accuracy is set as the accuracy of running KCFU for $15$ rounds. We experiment with the baseline method by replacing KCFU with a vanilla federated update which removes $\mathcal{L}_\text{compen}$ and updates with $\mathcal{L}_\text{client}$ (eq.~\eqref{eq:g-loss}) only. We also compare the results from mixing and not mixing unlabelled data in KCFU. 
As shown in Tab.~\ref{tab:KCFU}, KCFU can converge faster than vanilla update no matter mixed data are used, demonstrating the effect of our knowledge-compensatory design which borrows common knowledge from the global model. Mixing data in KCFU further boosts the efficiency. 
We further experiment by fixing ${\bm\Gamma(\bm{\tilde{x}'})}=\frac{1}{C}$, where $C$ is the number of classes, for $\mathcal{L}_\text{compen}$ (Eq.~\eqref{eq:compen-loss}). This means we distil knowledge from the downloaded global model without differentiation on all unlabelled data. Unsurprisingly, the performance is very poor. The accuracy reaches only $40.4\%$ with $10\%$ data and the setting aligned with Fig.~\ref{fig:experiments}(a).

\begin{table}[]
\caption{Number of rounds of different federated update ways to achieve the same accuracy as running KCFU runs for $15$ rounds.}
\centering
\label{tab:KCFU}
\begin{tabular}{|l|c|c|}
\hline
Method & CIFAR10 & CIFAR100 \\ \hline
Baseline & 35  & 39 \\ \hline 
KCFU w/o mix & 25 & 27 \\ \hline
\bf KCFU & \bf 15 & \bf 15 \\ \hline
\end{tabular}
\end{table}

\subsubsection{Local update without the balanced loss}
We use a balanced loss (Eq.~\eqref{eq:g-loss}) for local update of clients. This type of loss is usually the cherry on the top for standard federated learning. This is however not the case in our federated active learning problem. In Fig.~\ref{fig:experiments}(f), we show the results using balanced loss (Eq.~\eqref{eq:g-loss}) and a simple cross-entropy loss (simply replacing $n_{i,y}$ and $n_{i,c}$ in Eq.~\eqref{eq:g-loss} with $1$). 
We ran the experiment with two methods, our KAFAL and random sampling. From the experiment results, we can see that removing the balanced loss in local update disturbs or almost ruins the learning, a drop of $5\%$ to $10\%$ in performance occurs. Our KAFAL still outperforms random sampling, but the results are highly unstable. This is somewhat foreseeable since each client model starts with a very small amount of data in federated active learning. Despite our learning to intensify on specialized knowledge during sampling, it is still crucial to handle the imbalance of data during local client update using the balanced loss. 
 
\subsubsection{Knowledge specialization alternatives}
It is an interesting question whether other reweighting techniques can also help achieve knowledge specialization in federated active learning. Here we compare our method with two knowledge specialization alternatives, probability-level specialization and KL-Divergence-level specialization. Results and detailed analyses are presented in the supplementary (\ref{supp:a2}). 
The experimental results show that KAFAL outperforms both of the alternative methods. While probability-level specialization yields an acceptable outcome, KL-Divergence-level specialization fails to produce a reasonable result. One possible reason for this difference is that the probability-level specialization method, like our KAFAL, uses a moderate level of reweighting to adjust the results. In contrast, the KL-Divergence-level specialization method directly reweights the summation in the KL-Divergence calculation, potentially resulting in a level of reweighting that is too strong. 

\subsubsection{Different non-IID levels} 
We further explore federated active learning with the non-IID coefficient $\alpha=0.3$ and $\alpha=1$ on CIFAR10. We show results and detailed analysis in the supplementary (\ref{supp:a3}). 
A larger $\alpha$ value provides less non-IID distributions for clients, i.e., the distributions across different clients are more similar. 
Unsurprisingly, compared to our CIFAR10 with $\alpha=0.1$ results, the results are overall better for $\alpha=0.3$ and $\alpha=1$. 
Our KAFAL is still state-of-the-art for $\alpha=0.3$ and $\alpha=1$, but the margins between the results of KAFAL and the rest methods are relatively smaller. This experiment demonstrates that our KAFAL is more competitive with higher levels of non-IID. It validates that intensifying knowledge-specialized data in KAFAL can handle the non-IID distributed data in federated active learning.

\subsubsection{Different $\lambda$ values} 
{
The coefficient $\lambda$ in eq.~\eqref{eq:ks-prob-p}\eqref{eq:ks-prob-q} controls the knowledge-specialized level in KSAS. With larger values of $\lambda$, the clients intensify more on their specialized knowledge in active sampling. As we stated, we simply use $\lambda=1$ in our main experiments. Here we explore more values of $\lambda$ on CIFAR10 and show the results and detailed analysis in the supplementary (\ref{supp:a4}). For $\lambda$ of values $1$, $2$, and $3$, the difference is not significant. However, the results are clearly poorer for more extreme $\lambda$ values $0.1$ and $10$. Therefore, when applying KAFAL, the selection of $\lambda$ value can be flexible, but the chosen value should be neither too small nor too large.

}

\subsubsection{Learning with more decentralized clients} 
In previous sections, we explored federated active learning with $N=10$ clients. To better analyze the problem, we run experiments on CIFAR10 with $N=20$ and $N=100$. 
The labelled data amount still starts with $10\%$ of each training set, meaning that the local dataset on each client is smaller in size. 
The results and detailed analysis are shown in the supplementary (\ref{supp:a5}).
Compared with results from using $N=10$ clients, results for all methods reduce due to the smaller local datasets. 
Our KAFAL still outperforms the rest methods by a clear margin. 
\subsubsection{A smaller ratio of clients to update per round} 
We used $R=80\%$ in previous experiments. To test how our KAFAL performs with a smaller ratio of clients updated in each communication round, we use $R=40\%$, meaning that only $40\%$ of the clients are updated in each communication round. 
Surprisingly, our KAFAL performs even better using $R=40\%$ compared with using $R=80\%$, while results from the rest methods all drop. This is possibly because our KAFAL compensates for the knowledge of clients with the global model using KCFU along with actively sampling data by intensifying specialized knowledge using KSAS. The two together enable a faster convergence in global aggregations. 
Using $R=40\%$ means each client is trained less compared to using $R=80\%$ when the communication rounds $T$ is fixed. The rest methods which still actively sample harder data that are likely from less frequent classes cannot utilize these data in training with the smaller $R$ value. Although KCFU is also used for other methods for a fair comparison, it cannot be fully utilized without the knowledge-specialized intensification of KSAS. 
Detailed results and analyses are shown in the supplementary (\ref{supp:a6}). 
\begin{figure}[t]
    \centering
    \includegraphics[trim=0 350 0 0,width=\linewidth, page=33]{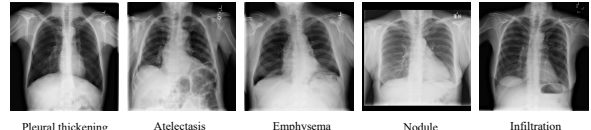}
    \caption{Selected images in NIH Chest X-Ray dataset. 
    }
    \label{fig:ncxr}
\end{figure}

\subsection{Medical Image Classification}
We further conduct experiments in a more realistic scenario of X-ray image classification using NIH Chest X-Ray dataset~\cite{wang2017hospital}. Some examples are shown in Fig.~\ref{fig:ncxr}. The task is to categorize thorax diseases using chest X-ray images. The dataset consists of more than 112k images of size $1024\times 1024$.  We follow the official training and testing splits. And we exclude images tagged with 'no findings'. The rest data have $14$ for different thorax diseases as labels. The training split includes $36024$ images and the testing split includes $15735$ images. We use ResNet-50~\cite{he2016deep} as the backbone of the clients and the global model. We still use $\alpha=0.1$ as the non-IID coefficient to distribute the client data. $5$ clients are used, and $80\%$ are selected for the update at each communication round. We start with $10\%$ labels and use $5\%$ of the whole dataset as the budget. The results are shown in the supplementary (\ref{supp:a7}). We compare with four baseline methods (Random, Core-Set, Entropy, and Margin) that the dataset can easily fit in considering the image size and model size. Our KAFAL still achieves state-of-the-art results on this dataset.

\section{Conclusion}
We have introduced a federated active learning paradigm which allows actively selecting the unlabelled data to efficiently learn a global model given a limited annotation budget in a decentralized learning process. We revealed that the main challenge in federated active learning is the mismatch between the active sampling goals of the global model on the server and each local client due to model differences caused by non-IID data distributions. This paper devised a Knowledge-Aware Federated Active Learning (KAFAL) method for federated active learning with non-IID data. KAFAL computes the discrepancies between client-server models with an intensification on each client's specialized knowledge. It is worth noting that the intensifying process is particularly important to achieve a powerful global model in the non-IID federated learning framework. Moreover, KAFAL also compensates for each client's ability in rare classes to handle data heterogeneity caused by non-IID data during federated updates.
Extensive experiments and analyses have validated the superiority of KAFAL over the state-of-the-art active learning methods under the federated active learning framework.

\clearpage
\renewcommand\thesubsection{\Alph{subsection}}

\section*{Supplementary}
In this supplementary, we provide additional experimental details and visualizations that provide further insights into the main content presented in the paper. Furthermore, we discuss the limitations of our work and highlight possible future directions for improvement.

\subsection{Additional Experimental Details and Visualizations}
In this section, we provide additional details of our experiments in the paper to further support our main content. We include visualizations of data distributions, hardware information, and detailed result numbers to help readers better understand our experimental setup. Furthermore, we present results from extra ablation studies and on additional datasets to demonstrate the effectiveness and robustness of our proposed KAFAL approach. Moreover, we introduce a visualization that illustrates the mismatch of active sampling goals between the global model and the client models in federated active learning, highlighting the importance of knowledge-specialized sampling for better performance. We also provide a detailed demonstration of the Knowledge-Specialized KL-Divergence using a toy example, to help readers better understand this key component of our approach.
\subsubsection{Experiment Details}
\label{supp:a1}
\begin{figure*}[t]
    \centering
    \includegraphics[trim= 0 300 0 0, width=0.95\textwidth, page=41]{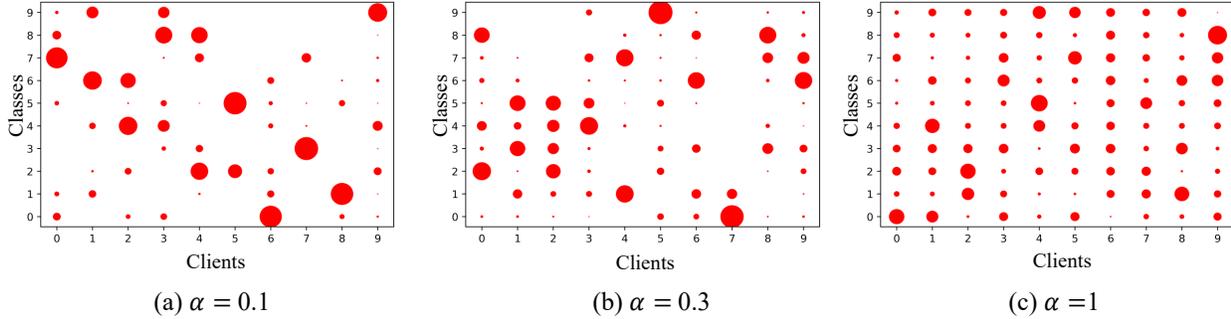}
    \caption{Illustration of CIFAR10 non-IID data distributions over clients with $\alpha=0.1$, $\alpha=0.3$, and $\alpha=1$. The $x$-axes represent the client names. The $y$-axes represent the class labels. The dot sizes represent the number of data. 
    }
    \label{fig:data}
\end{figure*}
\begin{figure*}[t]
    \centering
    \includegraphics[trim= 0 300 0 0, width=0.95\textwidth, page=61]{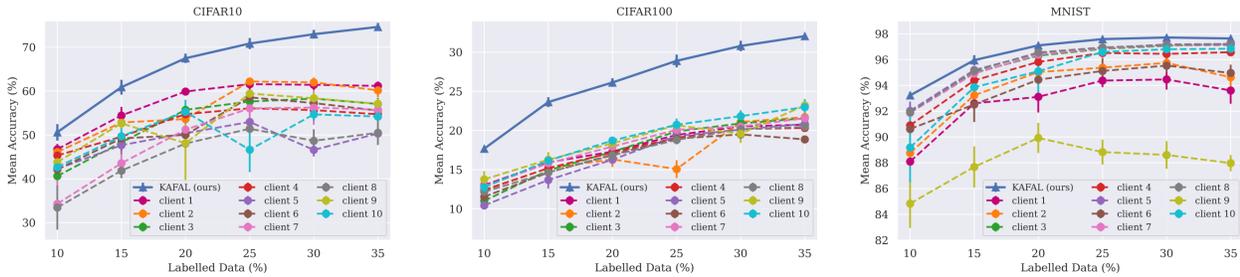}
    \caption{Evaluation on each client on CIFAR 10/100 and MNIST using KAFAL.
    }
    \label{fig:clientwise}
\end{figure*}

The experiments are conducted with one NVIDIA GeForce GTX 1080 Ti GPU. Each client is trained for $40$ epochs locally. The batch size is $128$ and the learning rate $\eta=0.1$. We run $T=50$ communication rounds before active sampling and evaluation.  ${\bm \beta}$ is sampled from $\text{Beta}(2,2)$ and $\nu=0.5$. 
We present the exact result numbers of our main results on CIFAR10 (Tab.~\ref{tab:cifar10}) and CIFAR100 (Tab.~\ref{tab:cifar100}). 
To present a complete picture of our KAFAL performance, we evaluate each client using the test set and show the results in Fig.~\ref{fig:clientwise}.
The non-IID data distributions used in our main results on CIFAR10 are show in Fig.~\ref{fig:data}(a).  

\setlength{\tabcolsep}{4.0pt}
\begin{table}[htb]
\caption{Detailed results on CIFAR10.}
\label{tab:cifar10}
\resizebox{\linewidth}{!}{
\begin{tabular}{|l|l|l|l|l|l|l|}
\hline
Method & 10\% & 15\% & 20\% & 25\% & 30\% & 35\% \\ \hline
Random & \multirow{8}{*}{50.60} & 54.29 &59.76 &62.85 &65.16 &66.52  \\ \cline{1-1} \cline{3-7} 
Core-Set & &58.98 &67.48&68.85&69.04&71.05\\ \cline{1-1} \cline{3-7} 
Entropy &  & 58.45 &65.76 &68.61 &69.59 & 71.68 \\ \cline{1-1} \cline{3-7} 
Margin &  &58.19 &63.50 & 66.75 &68.66 &71.13 \\ \cline{1-1} \cline{3-7} 
LL4AL &  &  57.48 &60.87 &63.79 &65.31 &66.94\\ \cline{1-1} \cline{3-7} 
QBC &  & 58.45 & 62.10 &65.81 &66.49 & 68.22\\ \cline{1-1} \cline{3-7} 
BADGE &  &57.46 &63.57 &67.39 &70.42 &71.67  \\ \cline{1-1}
\cline{3-7} 
Alfa-Mix &  &56.75&61.90&64.98&66.57&67.81 \\ \cline{1-1}
\cline{3-7} 
\bf KAFAL (ours) &  &60.88 & 67.47 & 70.82&72.94 &74.60\\ \hline
\end{tabular} 
}

\end{table}

\begin{table}[htb]
\caption{Detailed results on CIFAR100.}
\label{tab:cifar100}
\resizebox{\linewidth}{!}{
\begin{tabular}{|l|l|l|l|l|l|l|}%
\hline
Method & 10\% & 15\% & 20\% & 25\% & 30\% & 35\% \\ \hline
Random & \multirow{8}{*}{17.67} &20.10 &24.28 & 27.85 &29.21 &30.41  \\ \cline{1-1} \cline{3-7} 
Core-Set &  &22.78 &26.10 &28.49 &30.11 &30.86 \\ \cline{1-1} \cline{3-7} 
Entropy &  &20.79 &23.48 &26.41 &28.07 &30.01 \\ \cline{1-1} \cline{3-7} 
Margin &  & 22.65 &25.50 & 28.56 & 29.77 &30.88\\ \cline{1-1} \cline{3-7} 
LL4AL &  & 22.18 & 24.05 &27.14 &27.99 &28.41\\ \cline{1-1} \cline{3-7} 
QBC &  & 22.41 &24.86 &27.15 &29.95 &30.39\\ \cline{1-1} \cline{3-7} 
BADGE &  &22.93 &26.19 &28.61 &29.72 &31.26\\ \cline{1-1} \cline{3-7} Alfa-Mix &  &21.14 &24.54 &27.79&29.15&30.56\\ \cline{1-1}
\cline{3-7} 
\bf KAFAL (ours) &  & 23.63 & 26.13 & 28.89 & 30.79 &32.04 \\ \hline
\end{tabular}
}
\end{table}

\subsubsection{Knowledge Specialization Alternatives}
\label{supp:a2}
\begin{figure}[t]
    \centering
    \includegraphics[trim= 70 0 60 0 ,clip,width=0.5\textwidth, page=59]{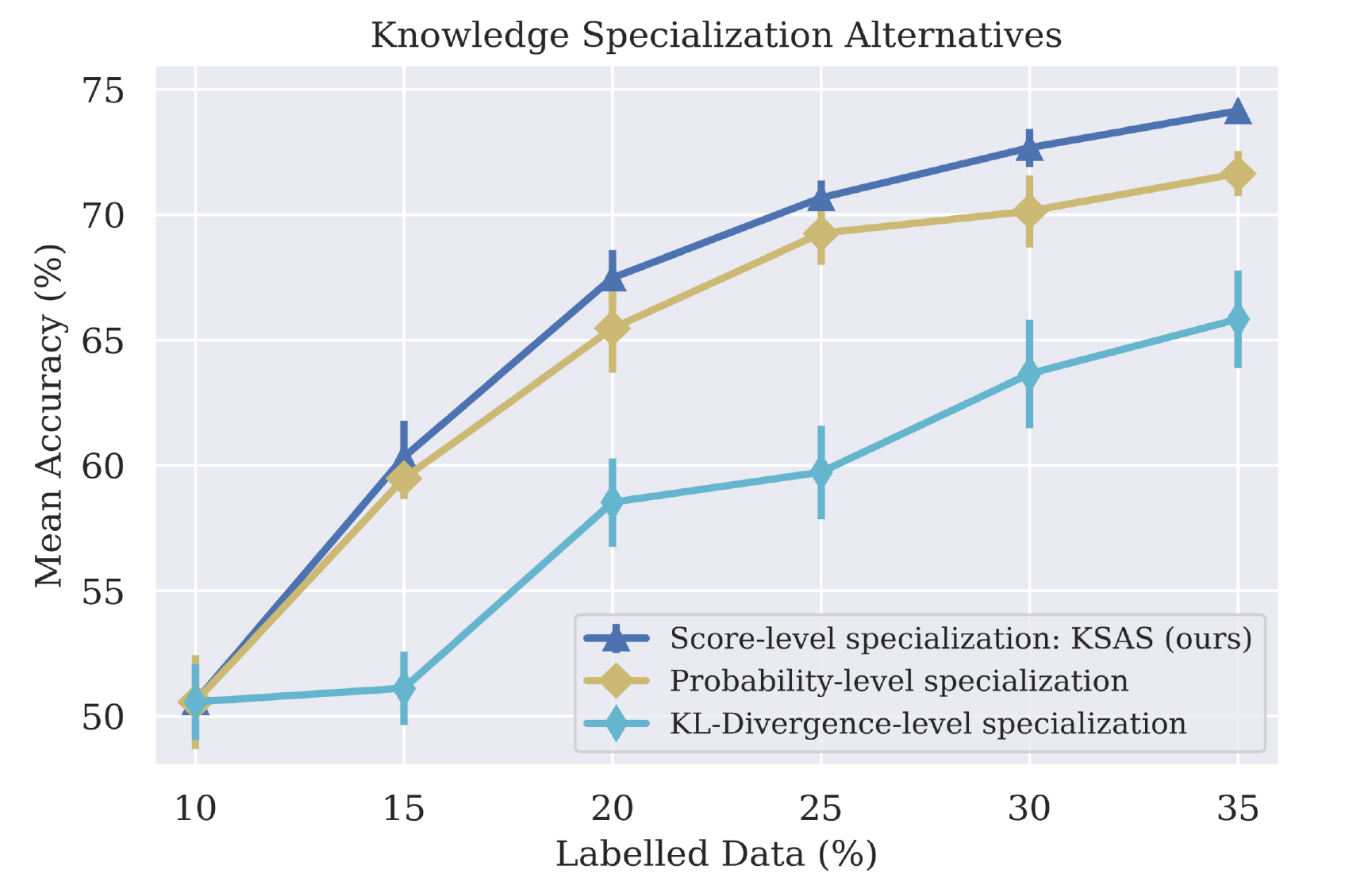}
    \caption{Results on CIFAR using different knowledge specialization techniques.}
    \label{fig:specialization}
\end{figure}
Given that knowledge specialization of KL-Divergence is achieved via score-level reweighting (as detailed in Eq.~(1)-(3) of the paper) in our KAFAL, an interesting question arises: Can other reweighting techniques also enable knowledge specialization in federated active learning? To answer this question, we compare our method with two knowledge specialization alternatives, namely probability-level specialization and KL-Divergence-level specialization.

To conduct probability-level specialization, we can rewrite Eq.~(1) as follows:
\begin{equation*}
P_{y}^i(\bm x) =  \frac{\text{exp}\Big(\nu_{i,y}^{\lambda}\cdot g_y({\bm x};{\bm\omega_i})\Big)}{\sum_{c\in\mathbb{C}}\text{exp}\Big(\nu_{i,c}^{\lambda}\cdot g_c({\bm x};{\bm\omega_i})\Big)},
\label{eq:prob-level-p}
\end{equation*}
where $\nu_{i,y}=\frac{n_{i,y}}{\sum_{c\in\mathbb{C}}n_{i,c}}$ is the normalized knowledge weight. Note that we did not normalize the knowledge weight in our score-level knowledge specialization (KAFAL) because it can be easily proved that the results are equivalent with or without normalization.
And similarly, Eq.~(2) is replaced with:
\begin{equation*}
Q_{y}^i(\bm x) =  \frac{\text{exp}\Big(\nu_{i,y}^{\lambda}\cdot g_y({\bm x};{\bm\Omega})\Big)}{\sum_{c\in\mathbb{C}}\text{exp}\Big(\nu_{i,c}^{\lambda}\cdot g_c({\bm x};{\bm\Omega})\Big)}.
\label{eq:prob-level-q}
\end{equation*}
This knowledge specialization alternative still involves the computation of the KL-Divergence as described in Eq.~(3).  This knowledge specialization alternative reweights the logits during the calculation of the predicted probability, hence the name.

To conduct KL-Divergence-level specialization, we replace Eq.~(3) with:
\begin{equation*}
\small
D^i(\bm x)=\sum_{y\in\mathbb{C}}\left[\nu_{i,y}^{\lambda}\cdot \Big( P_y^i(\bm x)\ln \frac{P_y^i(\bm x)}{Q_y^i(\bm x)} + Q_y^i(\bm x)\ln \frac{Q_y^i(\bm x)}{P_y^i(\bm x)} \Big)\right].
\label{eq:ks-kld-weight}
\end{equation*}
This knowledge specialization alternative reweights the sum while calculating KL-Divergence.

In Fig.~\ref{fig:specialization}, we present the results of the two alternatives as well as our KAFAL. The experimental results show that KAFAL outperforms both of the alternative methods. While probability-level specialization yields an acceptable outcome, KL-Divergence-level specialization fails to produce a reasonable result. One possible reason for this difference is that the probability-level specialization method, like our KAFAL, uses a moderate level of reweighting to adjust the results. In contrast, the KL-Divergence-level specialization method directly reweights the summation in the KL-Divergence calculation, potentially resulting in a stronger level of reweighting. Our score-level specialization approach may outperform probability-level specialization because reweighting the raw logits may not have a natural interpretation, whereas reweighting normalized results as in our KAFAL can be interpreted as adjusting the likelihood of the results. 

\subsubsection{Different Non-IID Levels} 
\label{supp:a3}
\begin{figure}[t]
    \centering
    \includegraphics[trim=10 0 600 0,clip,width=0.48\textwidth, page=42]{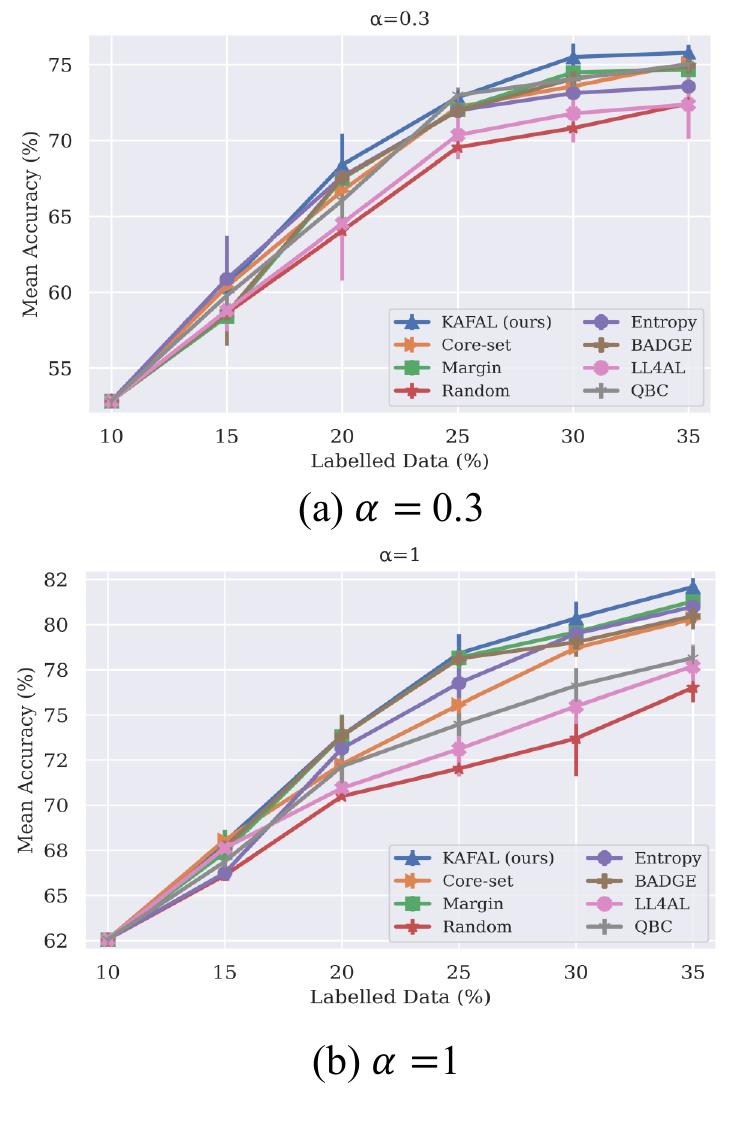}
    \caption{Results from using $\alpha=0.3$ and $\alpha=1$ for the non-IID coefficient on CIFAR10.
    }
    \label{fig:non-iid}
\end{figure}

We further explore federated active learning with the non-IID coefficient $\alpha=0.3$ and $\alpha=1$ on CIFAR10. The data distributions are shown in Fig.~\ref{fig:data}(b) and (c) respectively. We show the experiment results in Fig.~\ref{fig:non-iid}. A larger $\alpha$ value provides less non-IID distributions for clients, i.e., the distributions across different clients are more similar. Unsurprisingly, compared to our CIFAR10 with $\alpha=0.1$ results, the results are overall better for $\alpha=0.3$ and $\alpha=1$. Our KAFAL is still state-of-the-art, but the margins between the results of KAFAL and the rest methods are relatively smaller. This experiment demonstrates that our KAFAL is more competitive with higher levels of non-IID. It validates that intensifying knowledge-specialized data in KAFAL can handle the non-IID distributed data in federated active learning. The margins between Random and other methods become larger with larger $\alpha$ values, possibly because the mismatch problem in federated active learning becomes less significant with a lower level of non-IID in data. And the rest methods can benefit from the actively sampled data. 

\subsubsection{Different Values of $\lambda$ For Knowledge-Specialized Intensification} 
\label{supp:a4}
\begin{figure}[t]
    \centering
    \includegraphics[trim= 70 0 60 0 ,clip,width=0.5\textwidth, page=43]{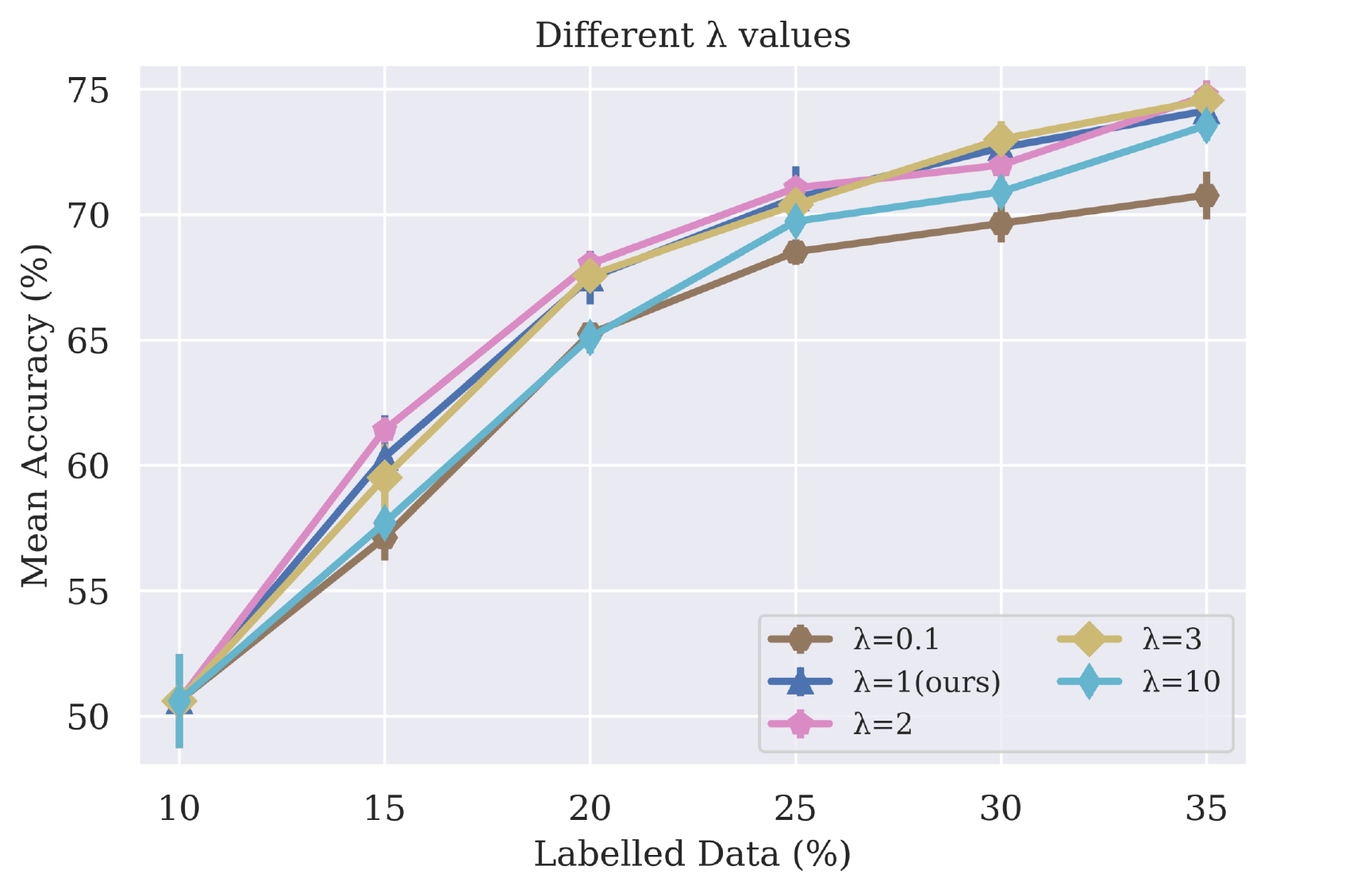}
    \caption{Results on CIFAR10 from using five different values of $\lambda$ for the intensification of specialized knowledge in federated active learning with non-IID data.}
    \label{fig:lambda}
\end{figure}
{
The coefficient $\lambda$ in eq. (1)(2) controls the knowledge-specialized level in KSAS. With larger values of $\lambda$, the clients intensify more on their specialized knowledge in active sampling. As we stated in the paper, we simply use $\lambda=1$ in our main experiments. Here we explore more values of $\lambda$ on CIFAR10 and show the results in Fig.~\ref{fig:lambda}. For $\lambda$ of values $1$, $2$, and $3$, the difference is not significant. However, for more extreme $\lambda$ values $0.1$ and $10$, the results are clearly poorer. Specifically, $\lambda=0.1$ produces the worst results of the five. 
When the $\lambda$ value approaches zero, the active sampling purely depends on the disagreement between the clients and the global model. The results gradually approach the results from using vanilla KL-Divergence in Subsec. 4.3.1 in the paper. When the $\lambda$ value goes to infinity, the active sampling process almost ignores the less frequent classes and tries to compute the disagreement solely based on the most common class (or classes) of each client. Therefore, when applying KAFAL, the $\lambda$ value should be neither too small nor too large.

}

\subsubsection{Learning With More Decentralized Clients} 
\label{supp:a5}
\begin{figure}[t]
    \centering
    \includegraphics[trim= 0 0 600 0 ,clip,width=0.5\textwidth, page=62]{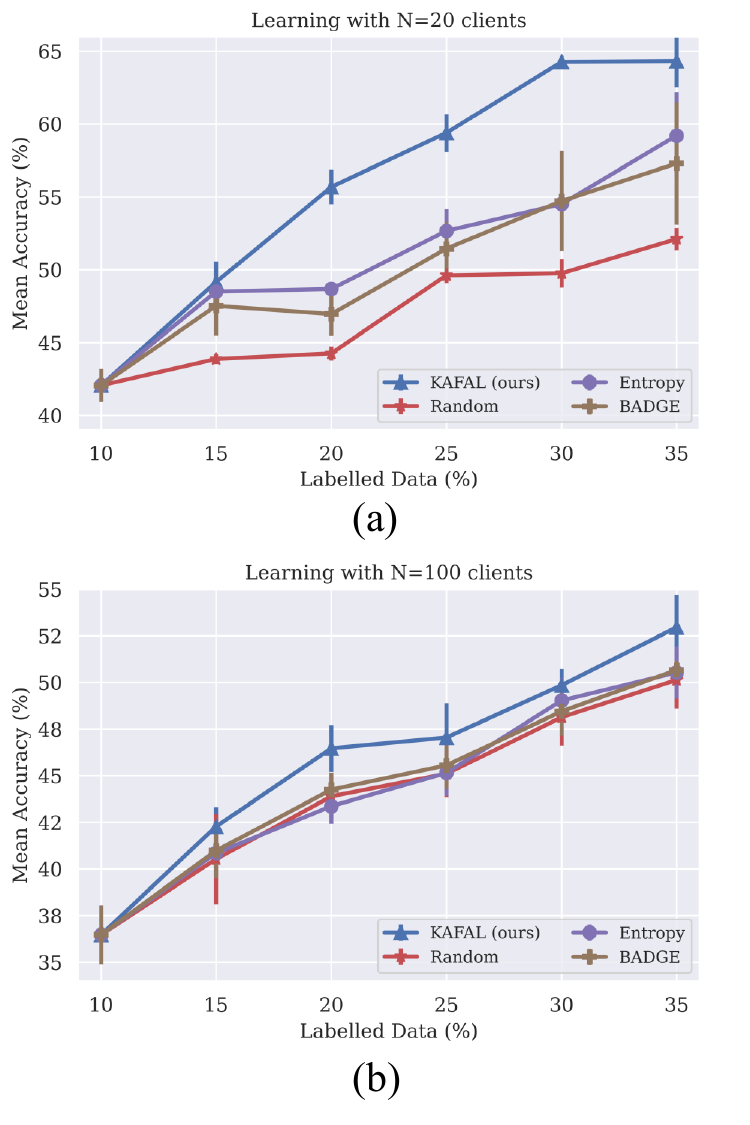}
    \caption{Results on CIFAR10 from using (a) $N=20$ and (b) $N=100$ clients in federated active learning with non-IID data.}
    \label{fig:20clients}
\end{figure}
In the paper, we explored federated active learning with $N=10$ clients. To better analyze the problem, we run experiments on CIFAR10 with $N=20$ and $N=100$ while keeping the rest setup the same. The labelled data amount still starts with $10\%$ of each local training set, meaning that with $N=20$ the data available for each client is half of that in the previous experiments, and with $N=100$ the data available for each client is only $\frac{1}{10}$ of that in the previous experiments. The results are shown in Fig.~\ref{fig:20clients}. Compared with the previous results from using $N=10$ clients, results for all methods reduce due to the smaller local datasets for both $N=20$ and $N=100$. Our KAFAL still outperforms the rest methods by a clear margin. This shows the superiority of our method when more decentralized clients are involved in federated active learning. The result lines are more jiggly compared with previous results, the possible reason is that the fewer labelled data and the $T=50$ communication rounds may not be enough for the convergence to be achieved. With $N=100$ clients, the margin is less significant compared to using $N=20$, this is possibly due to the extremely small local dataset size. Each local dataset starts with on average $500$ images and adds about $250$ images at each active round for $N=100$. This also explains why Entropy and BADGE generate similar results compared with Random. The limited training data lead to poor classification ability and deteriorates the credibility of the model statistics. 

\subsubsection{A Smaller Ratio of Clients to Update per Round} 
\label{supp:a6}
\begin{figure}[t]
    \centering
    \includegraphics[trim= 70 0 60 0 ,clip,width=0.5\textwidth, page=45]{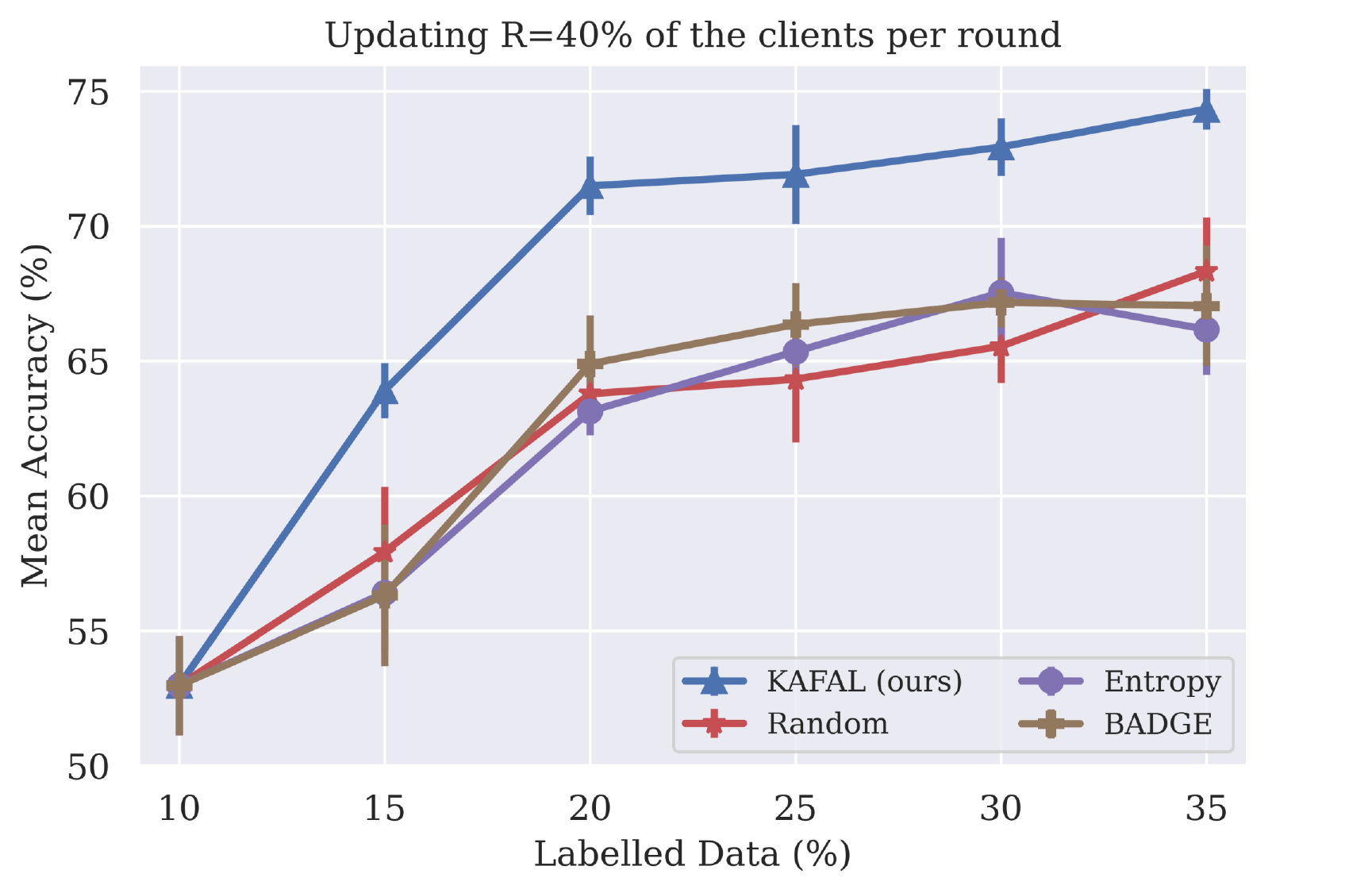}
    \caption{Results on CIFAR10 from updating $40\%$ of clients per communication round in federated active learning with non-IID data.}
    \label{fig:ratio0-4}
\end{figure}
We used $R=80\%$ in previous experiments. To test how our KAFAL performs with a smaller ratio of clients updated in each communication round, we use $R=40\%$ instead and present the results on CIFAR10 in Fig.~\ref{fig:ratio0-4}. All the rest setup is kept the same. $R=40\%$ means that only $40\%$ of the clients are updated in each communication round. Surprisingly, our KAFAL performs even better using $R=40\%$ compared with using $R=80\%$, while results from the rest methods all drop. This is possible because our KAFAL compensates for the knowledge of clients with the global model using KCFU along with actively sampling data by intensifying specialized knowledge using KSAS. The two together enable a faster convergence in global aggregations. Using $R=40\%$ means each client is trained less compared to using $R=80\%$ when the communication rounds $T$ is fixed. The rest methods which still actively sample harder data that are likely from less frequent classes cannot utilize these data in training with the smaller $R$ value. Although KCFU is also used for other methods for a fair comparison, it cannot be fully utilized without the knowledge-specialized intensification of KSAS.

\begin{figure}[t]
    \centering
    \includegraphics[trim=0 350 0 0,width=0.5\textwidth, page=33]{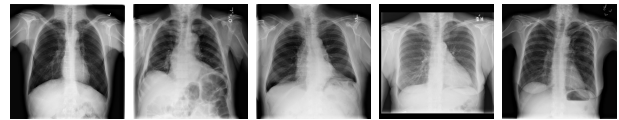}
    \caption{Selected images in NIH Chest X-Ray dataset. 
    }
    \label{fig:ncxr}
\end{figure}

\subsubsection{Medical Image Classification}
\label{supp:a7}
We further conduct experiments in a more realistic scenario of X-ray image classification using NIH Chest X-Ray dataset~\cite{wang2017hospital}. Some examples are shown in Fig.~\ref{fig:ncxr}. The task is to categorize thorax diseases using chest X-ray images. The dataset consists of more than 112k images of size $1024\times 1024$.  We follow the official training and testing splits. And we exclude images tagged with 'no findings'. The rest data have $14$ for different thorax diseases as labels. The training split includes $36024$ images and the testing split includes $15735$ images. We use ResNet-50~\cite{he2016deep} as the backbone of the clients and the global model. We still use $\alpha=0.1$ as the non-IID coefficient to distribute the client data. $5$ clients are used, and $80\%$ are selected for the update at each communication round. We start with $10\%$ labels and use $5\%$ of the whole dataset as the budget. We train for $2$ epochs in each communication round with learning rate $\eta=0.0005$ and run $5$ communication rounds before sampling. The mean AUC score is used to evaluate each method's performance. The results are presented in Tab.~\ref{tab:ncxr}. We compare with four baseline methods (Random, Core-Set, Entropy, and Margin) that the dataset can easily fit in considering the image size and model size. Our KAFAL still achieves state-of-the-art results on this dataset.

\begin{table}[htb]
\centering
\caption{mAUC scores on NIH Chest X-Ray dataset.}
\label{tab:ncxr}
\begin{tabular}{|l|c|c|c|}
\hline
Method & 10\% & 15\% & 20\% \\ \hline
Random  & \multirow{5}{*}{56.12} & 60.62 & 62.77 \\ \cline{1-1} \cline{3-4} 
Core-Set &  & 62.55 & 63.24 \\ \cline{1-1} \cline{3-4} 
Entropy &  & 63.13 & 63.80 \\ \cline{1-1} \cline{3-4} 
Margin &  & 60.19 & 62.81 \\ \cline{1-1} \cline{3-4} 
\bf KAFAL (ours) &  & 63.61 & 64.48 \\ \hline
\end{tabular}
\end{table}

\subsubsection{Results on MNIST} 
\label{supp:a8}
\begin{figure}[t]
    \centering
    \includegraphics[trim= 50 0 20 0 ,clip,width=0.5\textwidth, page=46]{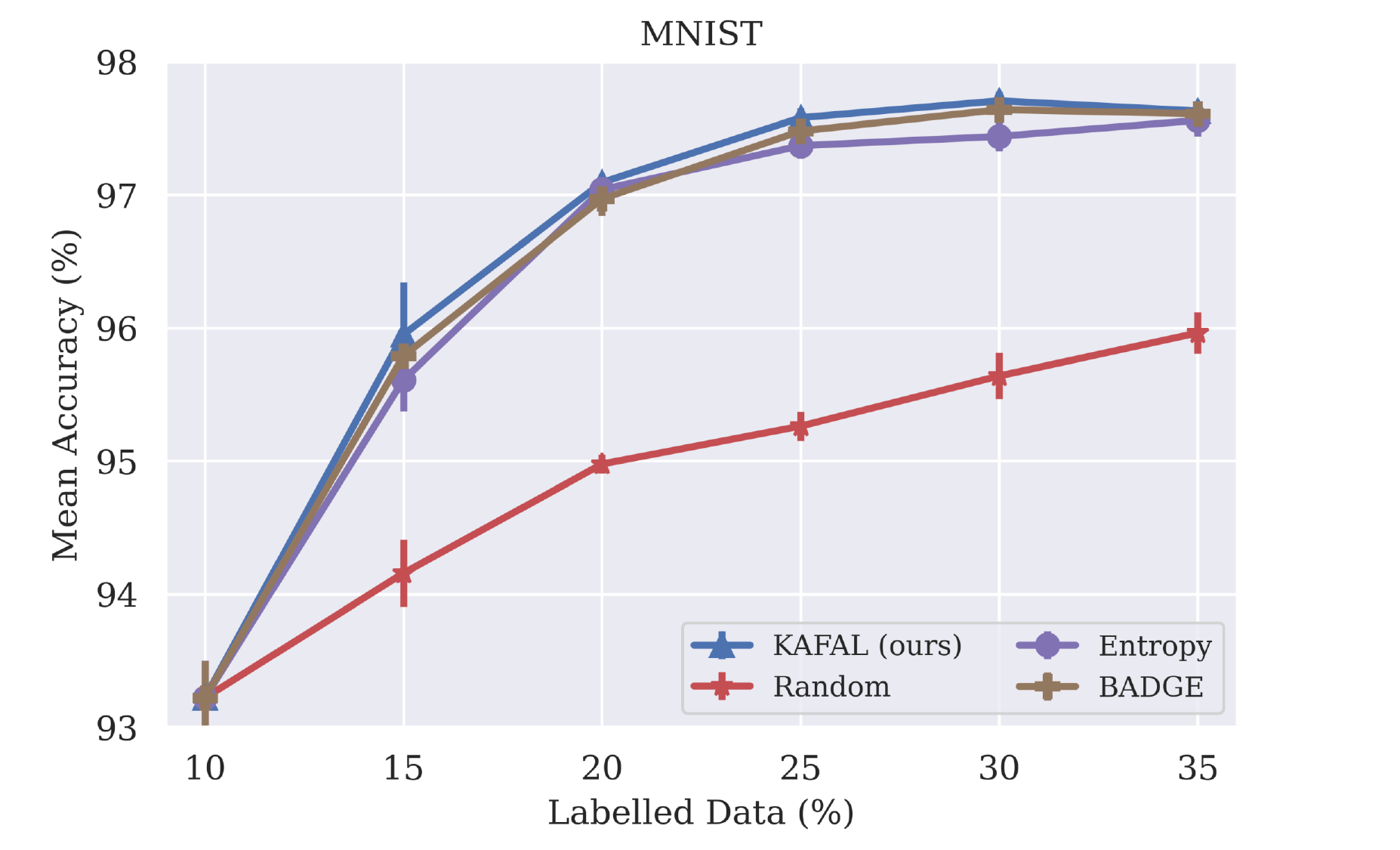}
    \caption{Results on MNIST in federated active learning with non-IID data.}
    \label{fig:mnist}
\end{figure}
We also run experiments on MNIST~\cite{lecun2010mnist}. MNIST is a $10$-class image dataset that contains handwritten images of $10$ digits. We use the MNIST 2NN proposed by McMahan et al.~\cite{mcmahan2017communication} as the clients' and the global model's architecture. We train for $10$ epochs each communication round and we repeat for $10$ communication rounds. We split the MNIST dataset with $\alpha=1$. The results are shown in Fig.~\ref{fig:mnist}. This is a fairly simple dataset, so all the results are quite high. But random is still far behind compared to the other methods. On this dataset, our KAFAL still outperforms the other methods, but with a quite small margin.

\subsubsection{Visualizing the Mismatch Problem} 
\label{supp:a9}
\begin{figure}[t]
    \centering
    \includegraphics[trim= 0 0 230 0 ,clip,width=0.45\textwidth, page=25]{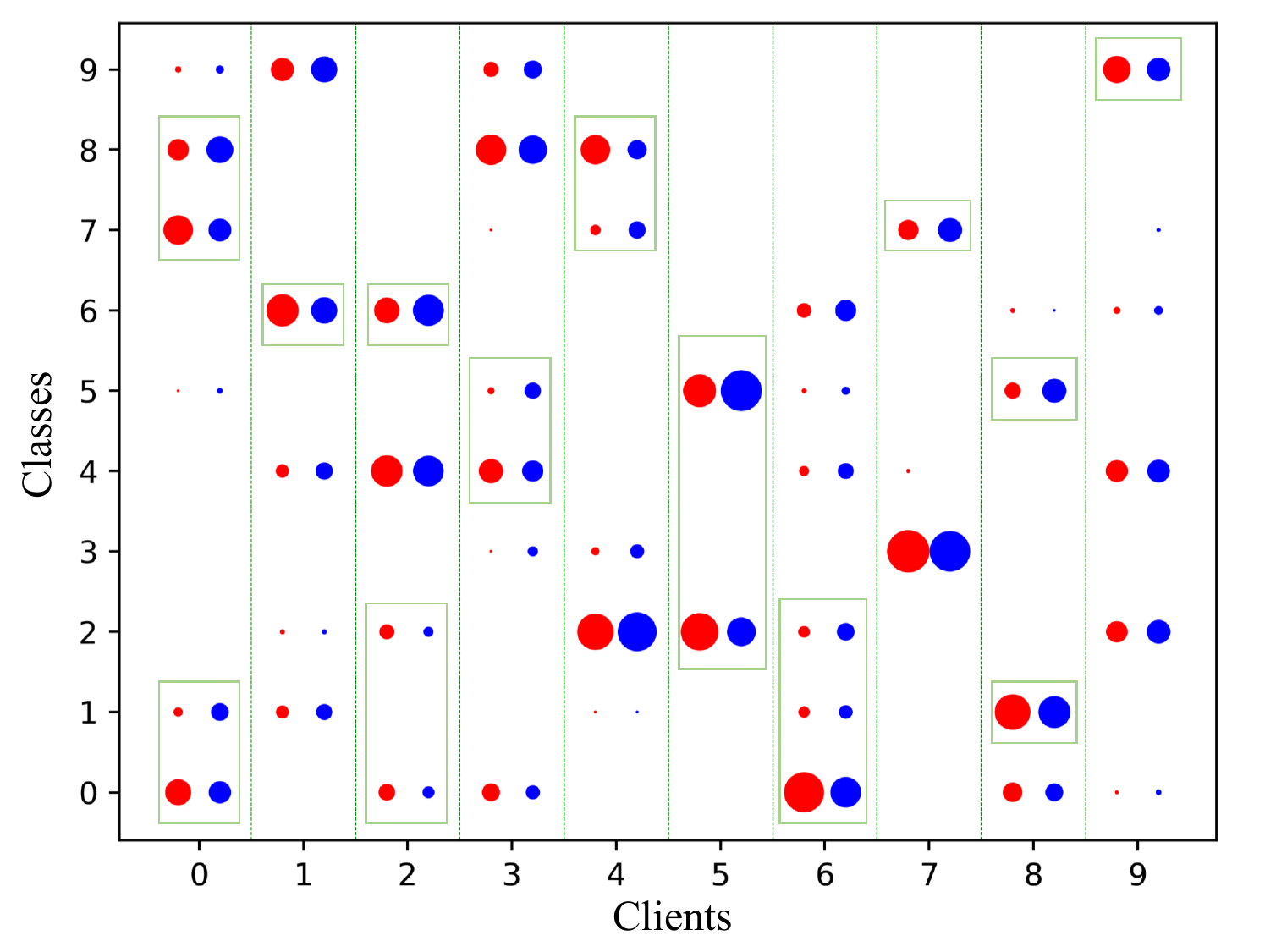}
    \caption{An example of the sampling goal mismatch between the global model and the clients. The green bounding boxes highlight class distributions that are clearly different for active sampling results using the global model (red) and active sampling results using the client model (blue). }
    \label{fig:difference}
\end{figure}
In the paper, we mentioned that the main challenge of federated active learning is the mismatch between the active sampling goal of the global model on the server and that of the asynchronous local clients. To demonstrate this problem with an experiment, we actively sample with the global model and the clients respectively and show the class distributions of the sampled data in Fig.~\ref{fig:difference}. We use the same sampling method Core-set for both the clients and the global model for a fair comparison. With the bounding boxes, we show the differences between the sampling results. The original data distributions on clients with $\alpha=0.1$ are shown in Fig.~\ref{fig:data}(a). Also, note that this figure only shows the class distributions. If we further consider specific data points within each class, the difference in sampled results will be more significant.

\subsubsection{Demonstration of Knowledge-Specialized KL-Divergence in a Toy Example with Details}
\label{supp:a10}
\begin{figure}[t]
    \centering
    \includegraphics[trim= 0 0 260 0 ,clip,width=0.5\textwidth, page=60]{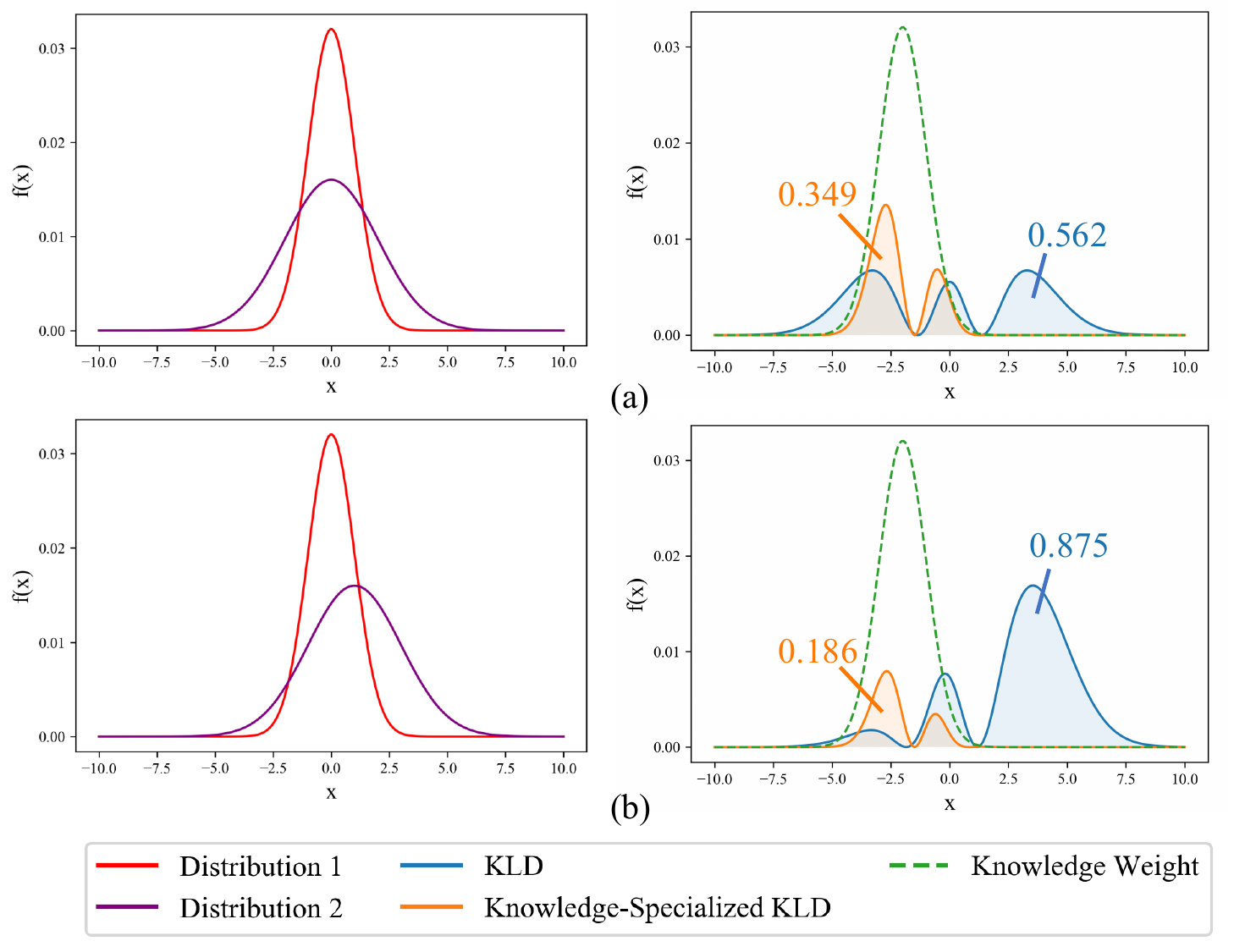}
    \caption{Illustration of how Knowledge-Specialized KL-Divergence intensifies specialized knowledge compared to standard KL-Divergence. 
    On the left, we show two distribution curves. On the right, the blue and orange lines integrate to be KL-Divergence and the knowledge-specialized KL-Divergence computed from the left distributions. The blue and orange numbers show the integrated areas of the blue and orange curves in each image, respectively. }
    \label{fig:toy}
\end{figure}
To better visualize how Knowledge-Specialized KL-Divergence intensifies specialized knowledge compared to KL-Divergence, we use continuous distributions to simulate model predictions and compute the divergences (Fig.~\ref{fig:toy}). Note that the knowledge weight curves serve as a continuous version of our knowledge weights. In the figure, we present the distribution curves on the left and the corresponding KL-Divergence and Knowledge-Specialized KL-Divergence curves on the right, which have been calculated accordingly. The KL-Divergence curve is formulated as:
\begin{equation*}
\small
p(x)\ln \frac{p(x)}{q(x)} + q(x)\ln \frac{q(x)}{p(x)} ,
\label{eq:kld-curve}
\end{equation*}
where $p(x)$ and $q(x)$ are the two distribution functions (presented on the left of Fig.~\ref{fig:toy}). The KL-Divergence value is obtained by integrating this function with respect to $x$. The Knowledge-Specialized KL-Divergence curve is formulated as:
\begin{equation*}
\small
p_w(x)\ln \frac{p_w(x)}{q_w(x)} + q_w(x)\ln \frac{q_w(x)}{p_w(x)},
\label{eq:kskld-curve}
\end{equation*}
where $p_w(x)= \frac{w(x) \cdot p(x)}{Z_p}$ and $q_w(x)=\frac{w(x) \cdot q(x)}{Z_q}$. The normalization constants $Z_p = \int p(x)\cdot w(x)\text{d}x$ and $Z_q = \int q(x)\cdot w(x)\text{d}x$. The weight curve $w(x)$ is shown with green dashed lines in the figure. The Knowledge-Specialized KL-Divergence value is obtained by integrating this function with respect to $x$. 
The right-hand side of Fig.~\ref{fig:toy} can be viewed as global-local discrepancies from two different inputs on the same client model since the KL-Divergence values are different and the knowledge weights are the same. On the left-hand side, distributions 1 and 2 simulate the outputs of the client model and the global model.
Notably, while (a) has a smaller KL-Divergence, its Knowledge-Specialized KL-Divergence is larger, suggesting it is less likely to be sampled than (b) if KL-Divergence is the sampling criterion. However, using our proposed Knowledge-Specialized KL-Divergence, (a) is more likely to be sampled than (b). This difference in sampling results is due to the knowledge weight, which intensifies the client's specialized knowledge while dampening the contribution of unfamiliar knowledge. Importantly, in (a), more of the model difference arises from specialized knowledge (as indicated by the peak area of the knowledge weight) compared to (b).

\subsection{Limitations and Future Work}
Our federated active learning paradigm KAFAL includes KSAS, a novel active sampling method to sample informative data using intensified discrepancies between the server and clients based on the specialized knowledge of each client, and KCFU, a federated update method to deal with data heterogeneity by compensating weak classes with the help from the global model. Although the experimental results demonstrate that KAFAL can perform well on the federated active learning task, we also want to highlight the potential drawbacks of this method. In KSAS, the specialized knowledge is extracted based on the class distributions of labelled local data. We may explore other ways to find a more comprehensive solution to represent the specialized knowledge, either, possibly not only considering the class distributions but also taking the training dynamics into account. In KCFU, the compensation is achieved through sampling the unlabelled data and then weighting them using the class distributions. Unfortunately, the data from weak classes may not be enough even though we include the unlabelled data. We may utilize the data generation techniques to generate more weak-class data for better knowledge compensation in the future. 
In addition to the potential drawbacks mentioned, another area for future work is to extend KAFAL to handle the case of long-tailed distribution in the federated active learning setting. In a long-tailed scenario, the local data can distribute globally long-tailed with some classes being rare for all clients. To consider active learning in such a scenario, additional resampling techniques and an improved version of knowledge-specialized KL-Divergence that takes the long-tailed distribution into account need to be included. 

{\small
\bibliographystyle{ieee_fullname}
\bibliography{kafalbib}

\begin{thebibliography}{10}\itemsep=-1pt

\bibitem{ahn2022federated}
Jin-Hyun Ahn, Kyungsang Kim, Jeongwan Koh, and Quanzheng Li.
\newblock Federated active learning (f-al): an efficient annotation strategy
  for federated learning.
\newblock {\em arXiv preprint arXiv:2202.00195}, 2022.

\bibitem{ash2020deep}
Jordan~T Ash, Chicheng Zhang, Akshay Krishnamurthy, John Langford, and Alekh
  Agarwal.
\newblock Deep batch active learning by diverse, uncertain gradient lower
  bounds.
\newblock In {\em ICLR}, 2020.

\bibitem{beluch2018power}
William~H Beluch, Tim Genewein, Andreas N{\"u}rnberger, and Jan~M K{\"o}hler.
\newblock The power of ensembles for active learning in image classification.
\newblock In {\em CVPR}, 2018.

\bibitem{chen2021bridging}
Hong-You Chen and Wei-Lun Chao.
\newblock On bridging generic and personalized federated learning for image
  classification.
\newblock In {\em International Conference on Learning Representations}, 2021.

\bibitem{chen2020autodal}
Xu Chen and Brett Wujek.
\newblock Autodal: Distributed active learning with automatic hyperparameter
  selection.
\newblock In {\em Proceedings of the AAAI conference on artificial
  intelligence}, volume~34, pages 3537--3544, 2020.

\bibitem{cortes2019active}
Corinna Cortes, Giulia DeSalvo, Mehryar Mohri, Ningshan Zhang, and Claudio
  Gentile.
\newblock Active learning with disagreement graphs.
\newblock In {\em ICML}, 2019.

\bibitem{dagan1995committee}
Ido Dagan and Sean~P Engelson.
\newblock Committee-based sampling for training probabilistic classifiers.
\newblock In {\em Machine Learning Proceedings 1995}, pages 150--157. Elsevier,
  1995.

\bibitem{ebrahimi2020minimax}
Sayna Ebrahimi, William Gan, Dian Chen, Giscard Biamby, Kamyar Salahi, Michael
  Laielli, Shizhan Zhu, and Trevor Darrell.
\newblock Minimax active learning.
\newblock {\em arXiv preprint arXiv:2012.10467}, 2020.

\bibitem{freund1993information}
Yoav Freund, H~Sebastian Seung, Eli Shamir, and Naftali Tishby.
\newblock Information, prediction, and query by committee.
\newblock In {\em NIPS}, 1993.

\bibitem{fu2021transferable}
Bo Fu, Zhangjie Cao, Jianmin Wang, and Mingsheng Long.
\newblock Transferable query selection for active domain adaptation.
\newblock In {\em Proceedings of the IEEE/CVF Conference on Computer Vision and
  Pattern Recognition}, pages 7272--7281, 2021.

\bibitem{gao2020consistency}
Mingfei Gao, Zizhao Zhang, Guo Yu, Sercan~{\"O} Ar{\i}k, Larry~S Davis, and
  Tomas Pfister.
\newblock Consistency-based semi-supervised active learning: Towards minimizing
  labeling cost.
\newblock In {\em European Conference on Computer Vision}, pages 510--526.
  Springer, 2020.

\bibitem{goetz2019active}
Jack Goetz, Kshitiz Malik, Duc Bui, Seungwhan Moon, Honglei Liu, and Anuj
  Kumar.
\newblock Active federated learning.
\newblock {\em arXiv preprint arXiv:1909.12641}, 2019.

\bibitem{gong2021ensemble}
Xuan Gong, Abhishek Sharma, Srikrishna Karanam, Ziyan Wu, Terrence Chen, David
  Doermann, and Arun Innanje.
\newblock Ensemble attention distillation for privacy-preserving federated
  learning.
\newblock In {\em Proceedings of the IEEE/CVF International Conference on
  Computer Vision}, pages 15076--15086, 2021.

\bibitem{gudovskiy2020deep}
Denis Gudovskiy, Alec Hodgkinson, Takuya Yamaguchi, and Sotaro Tsukizawa.
\newblock Deep active learning for biased datasets via fisher kernel
  self-supervision.
\newblock In {\em Proceedings of the IEEE/CVF Conference on Computer Vision and
  Pattern Recognition}, pages 9041--9049, 2020.

\bibitem{he2016deep}
Kaiming He, Xiangyu Zhang, Shaoqing Ren, and Jian Sun.
\newblock Deep residual learning for image recognition.
\newblock In {\em Proceedings of the IEEE conference on computer vision and
  pattern recognition}, pages 770--778, 2016.

\bibitem{he2016identity}
Kaiming He, Xiangyu Zhang, Shaoqing Ren, and Jian Sun.
\newblock Identity mappings in deep residual networks.
\newblock In {\em European conference on computer vision}, pages 630--645.
  Springer, 2016.

\bibitem{hsieh2020non}
Kevin Hsieh, Amar Phanishayee, Onur Mutlu, and Phillip Gibbons.
\newblock The non-iid data quagmire of decentralized machine learning.
\newblock In {\em International Conference on Machine Learning}, pages
  4387--4398. PMLR, 2020.

\bibitem{hsu2019measuring}
Tzu-Ming~Harry Hsu, Hang Qi, and Matthew Brown.
\newblock Measuring the effects of non-identical data distribution for
  federated visual classification.
\newblock {\em arXiv preprint arXiv:1909.06335}, 2019.

\bibitem{hsu2020federated}
Tzu-Ming~Harry Hsu, Hang Qi, and Matthew Brown.
\newblock Federated visual classification with real-world data distribution.
\newblock In {\em European Conference on Computer Vision}, pages 76--92.
  Springer, 2020.

\bibitem{huang2021semi}
Siyu Huang, Tianyang Wang, Haoyi Xiong, Jun Huan, and Dejing Dou.
\newblock Semi-supervised active learning with temporal output discrepancy.
\newblock In {\em Proceedings of the IEEE/CVF International Conference on
  Computer Vision}, pages 3447--3456, 2021.

\bibitem{jeffreys1939theory}
Harold Jeffreys.
\newblock Theory of probability.
\newblock 1939.

\bibitem{jeong2020federated}
Wonyong Jeong, Jaehong Yoon, Eunho Yang, and Sung~Ju Hwang.
\newblock Federated semi-supervised learning with inter-client consistency \&
  disjoint learning.
\newblock In {\em International Conference on Learning Representations}, 2020.

\bibitem{kim2021task}
Kwanyoung Kim, Dongwon Park, Kwang~In Kim, and Se~Young Chun.
\newblock Task-aware variational adversarial active learning.
\newblock In {\em CVPR}, 2021.

\bibitem{kimlg}
SangMook Kim, SangMin Bae, Se-Young Yun, and Hwanjun Song.
\newblock Lg-fal: Federated active learning strategy using local and global
  models.

\bibitem{konevcny2016federated}
Jakub Kone{\v{c}}n{\`y}, H~Brendan McMahan, Daniel Ramage, and Peter
  Richt{\'a}rik.
\newblock Federated optimization: Distributed machine learning for on-device
  intelligence.
\newblock {\em arXiv preprint arXiv:1610.02527}, 2016.

\bibitem{krizhevsky2009learning}
Alex Krizhevsky, Geoffrey Hinton, et~al.
\newblock Learning multiple layers of features from tiny images.
\newblock 2009.

\bibitem{kullback1951information}
Solomon Kullback and Richard~A Leibler.
\newblock On information and sufficiency.
\newblock {\em The annals of mathematical statistics}, 22(1):79--86, 1951.

\bibitem{lecun2010mnist}
Yann LeCun, Corinna Cortes, and CJ Burges.
\newblock Mnist handwritten digit database.
\newblock {\em ATT Labs [Online]. Available: http://yann.lecun.com/exdb/mnist},
  2, 2010.

\bibitem{li2020federated}
Tian Li, Anit~Kumar Sahu, Manzil Zaheer, Maziar Sanjabi, Ameet Talwalkar, and
  Virginia Smith.
\newblock Federated optimization in heterogeneous networks.
\newblock {\em Proceedings of Machine Learning and Systems}, 2:429--450, 2020.

\bibitem{lin2020ensemble}
Tao Lin, Lingjing Kong, Sebastian~U Stich, and Martin Jaggi.
\newblock Ensemble distillation for robust model fusion in federated learning.
\newblock {\em Advances in Neural Information Processing Systems},
  33:2351--2363, 2020.

\bibitem{lu2021unsupervised}
Nan Lu, Zhao Wang, Xiaoxiao Li, Gang Niu, Qi Dou, and Masashi Sugiyama.
\newblock Unsupervised federated learning is possible.
\newblock In {\em International Conference on Learning Representations}, 2021.

\bibitem{ma2021active}
Xinhong Ma, Junyu Gao, and Changsheng Xu.
\newblock Active universal domain adaptation.
\newblock In {\em Proceedings of the IEEE/CVF International Conference on
  Computer Vision}, pages 8968--8977, 2021.

\bibitem{mahmood2021low}
Rafid Mahmood, Sanja Fidler, and Marc~T Law.
\newblock Low budget active learning via wasserstein distance: An integer
  programming approach.
\newblock {\em arXiv preprint arXiv:2106.02968}, 2021.

\bibitem{marfoq2021federated}
Othmane Marfoq, Giovanni Neglia, Aur{\'e}lien Bellet, Laetitia Kameni, and
  Richard Vidal.
\newblock Federated multi-task learning under a mixture of distributions.
\newblock {\em Advances in Neural Information Processing Systems}, 34, 2021.

\bibitem{mcmahan2017communication}
Brendan McMahan, Eider Moore, Daniel Ramage, Seth Hampson, and Blaise~Aguera y
  Arcas.
\newblock Communication-efficient learning of deep networks from decentralized
  data.
\newblock In {\em Artificial intelligence and statistics}, pages 1273--1282.
  PMLR, 2017.

\bibitem{melville2004diverse}
Prem Melville and Raymond~J Mooney.
\newblock Diverse ensembles for active learning.
\newblock In {\em ICML}, page~74, 2004.

\bibitem{mohri2019agnostic}
Mehryar Mohri, Gary Sivek, and Ananda~Theertha Suresh.
\newblock Agnostic federated learning.
\newblock In {\em International Conference on Machine Learning}, pages
  4615--4625. PMLR, 2019.

\bibitem{parvaneh2022active}
Amin Parvaneh, Ehsan Abbasnejad, Damien Teney, Gholamreza~Reza Haffari, Anton
  van~den Hengel, and Javen~Qinfeng Shi.
\newblock Active learning by feature mixing.
\newblock In {\em Proceedings of the IEEE/CVF Conference on Computer Vision and
  Pattern Recognition}, pages 12237--12246, 2022.

\bibitem{paszke2017automatic}
Adam Paszke, Sam Gross, Soumith Chintala, Gregory Chanan, Edward Yang, Zachary
  DeVito, Zeming Lin, Alban Desmaison, Luca Antiga, and Adam Lerer.
\newblock Automatic differentiation in pytorch.
\newblock 2017.

\bibitem{peng2019federated}
Xingchao Peng, Zijun Huang, Yizhe Zhu, and Kate Saenko.
\newblock Federated adversarial domain adaptation.
\newblock In {\em International Conference on Learning Representations}, 2019.

\bibitem{ren2020balanced}
Jiawei Ren, Cunjun Yu, Xiao Ma, Haiyu Zhao, Shuai Yi, et~al.
\newblock Balanced meta-softmax for long-tailed visual recognition.
\newblock {\em Advances in Neural Information Processing Systems},
  33:4175--4186, 2020.

\bibitem{sener2017active}
Ozan Sener and Silvio Savarese.
\newblock Active learning for convolutional neural networks: A core-set
  approach.
\newblock {\em arXiv preprint arXiv:1708.00489}, 2017.

\bibitem{seung1992query}
H~Sebastian Seung, Manfred Opper, and Haim Sompolinsky.
\newblock Query by committee.
\newblock In {\em Proceedings of the fifth annual workshop on Computational
  learning theory}, pages 287--294, 1992.

\bibitem{sinha2019variational}
Samarth Sinha, Sayna Ebrahimi, and Trevor Darrell.
\newblock Variational adversarial active learning.
\newblock In {\em ICCV}, 2019.

\bibitem{wangdual}
Shuo Wang, Yuexiang Li, Kai Ma, Ruhui Ma, Haibing Guan, and Yefeng Zheng.
\newblock Dual adversarial network for deep active learning.
\newblock 2020.

\bibitem{wang2017hospital}
Xiaosong Wang, Yifan Peng, Le Lu, Zhiyong Lu, M Bagheri, and R Summers.
\newblock Hospital-scale chest x-ray database and benchmarks on
  weakly-supervised classification and localization of common thorax diseases.
\newblock In {\em IEEE CVPR}, volume~7, 2017.

\bibitem{wang2021federated}
Zhiguo Wang, Xintong Wang, Ruoyu Sun, and Tsung-Hui Chang.
\newblock Federated semi-supervised learning with class distribution mismatch.
\newblock {\em arXiv preprint arXiv:2111.00010}, 2021.

\bibitem{yao2022federated}
Chun-Han Yao, Boqing Gong, Hang Qi, Yin Cui, Yukun Zhu, and Ming-Hsuan Yang.
\newblock Federated multi-target domain adaptation.
\newblock In {\em Proceedings of the IEEE/CVF Winter Conference on Applications
  of Computer Vision}, pages 1424--1433, 2022.

\bibitem{yoo2019learning}
Donggeun Yoo and In~So Kweon.
\newblock Learning loss for active learning.
\newblock In {\em CVPR}, 2019.

\bibitem{yoon2021federated}
Jaehong Yoon, Wonyong Jeong, Giwoong Lee, Eunho Yang, and Sung~Ju Hwang.
\newblock Federated continual learning with weighted inter-client transfer.
\newblock In {\em International Conference on Machine Learning}, pages
  12073--12086. PMLR, 2021.

\bibitem{zhang2020state}
Beichen Zhang, Liang Li, Shijie Yang, Shuhui Wang, Zheng-Jun Zha, and Qingming
  Huang.
\newblock State-relabeling adversarial active learning.
\newblock In {\em CVPR}, 2020.

\bibitem{zhao2018federated}
Yue Zhao, Meng Li, Liangzhen Lai, Naveen Suda, Damon Civin, and Vikas Chandra.
\newblock Federated learning with non-iid data.
\newblock {\em arXiv preprint arXiv:1806.00582}, 2018.

\end{thebibliography}
}

\end{document}